\documentclass[conference]{IEEEtran}
\usepackage{amsmath,amssymb,amsfonts}
\usepackage{graphicx}
\usepackage{booktabs}
\usepackage{multirow}
\usepackage{xcolor}
\usepackage[colorlinks=true,linkcolor=black,citecolor=black,urlcolor=black]{hyperref}

\newcommand{\method}{Leap-VLA}
\newcommand{\tslow}{$f_{\mathrm{slow}}$}
\newcommand{\tfast}{$f_{\mathrm{fast}}$}

\title{Leap-VLA: High-Rate Robot Control\\
from a Low-Rate VLM Backbone}
\author{\IEEEauthorblockN{Lin Yang\textsuperscript{*},
Yunjie Shen, Zheyu Li, Zhenjie Tian, Yongli Wang, Dongze Huang,\\
Weiqi Wang, Yucheng Zhu, Zhaoquan Jiang, Guofeng Xie, Peigen Fang}
\IEEEauthorblockA{Embodied Algorithm Group, Automation PBU, Leapmotor.\\
\textsuperscript{*}Corresponding author: Lin Yang.}}

\begin{document}
\maketitle

% ============================================================================
\begin{abstract}
Vision-language-action (VLA) models re-encode the full observation through a
multi-billion-parameter vision--language model (VLM) at every control step,
even though scene semantics evolve at roughly 1\,Hz while reactive control
requires 20--50\,Hz. We present \method{}, a VLA that decouples the two: a
VLM runs as a low-frequency \emph{semantic anchor} whose \emph{cached
context} is reused across control steps, while a flow-matching action expert
operates at high frequency, consuming the cached semantics with a fresh,
low-latency visual stream from a compact encoder pretrained on robot video
with masked reconstruction, optical flow, and depth. The \emph{measured} age
$\tau$ of the cache conditions the expert through adaptive layer
normalization. Reusing the cache makes the policy a function of its age, and
training covers that function explicitly: cache ages sampled from real
episode history, plus an age-ramped pixel corruption that erases the stale
snapshot the way deployed staleness corrupts content. On the asynchronous
deployment stack, the recipe closes most of the staleness gap---on
LIBERO-Spatial, 61.4\%$\to$95.5\% at a 2\,s semantic refresh (95.9\% at
1\,s, within 1.6 points of the frequency-matched anchor); averaged over the
four LIBERO suites, the deployed policy holds 95.2\% at a 1\,s refresh and
94.5\% at 2\,s against a 96.3\% matched anchor. A forward-only
chunk-boundary correction (\emph{soft-nudge}), an autograd-free
reformulation of real-time-chunking guidance, carries per-control-step
re-planning without loss (97.1\%) at zero added latency. The same lens
exposes a second defect: under an instruction--scene mismatch the policy
keeps manipulating the scene, and a chain-of-thought verdict gate trained on
zero-cost negative episodes freezes execution exactly on mismatched
instructions without hurting feasible success.
\end{abstract}

% ============================================================================
\section{Introduction}
Vision-language-action models map camera observations and
language instructions directly to robot actions, and recent systems built on
pretrained VLMs have demonstrated strong generalization across tasks,
objects, and scenes~\cite{rt2,openvla,pi0,pi05}. A structural property shared
by essentially all of these systems is that semantic understanding and action
generation run \emph{in lockstep}: at every re-planning step the full
observation is re-encoded by the VLM backbone, and only then does the action
head run.

This coupling is mismatched to the statistics of manipulation. Scene
semantics---which objects are present, where they roughly are, what the task
requires---change at about 1\,Hz and are often quasi-static for seconds at a
time. Reactive control, in contrast, must respond to proprioceptive drift,
contact events, and short-horizon object motion at 20--50\,Hz. Running a
multi-billion-parameter VLM at control frequency wastes the vast majority of
its compute on observations that are semantically identical to the previous
step, and it caps the re-planning rate at whatever the slow system affords.
The mismatch becomes prohibitive as backbones scale or as deployment moves to
edge hardware.

The obvious fix---reuse the VLM's processed features across control steps
and re-run only the action head---fails in two specific ways. First, the
cached prefix goes \emph{stale}: it describes a past state of the world, and
in current VLA designs it typically also embeds the \emph{robot state as
text} in the prompt, so the cache actively contradicts the fresh observation
as soon as the robot moves. We measure this on the deployed asynchronous
stack: a policy fine-tuned at the semantic refresh rate reaches 97.5\%, but
the same policy trained synchronously and then deployed with the semantic
loop slowed to a 2\,s refresh collapses to 61.4\% on LIBERO-Spatial
(Sec.~\ref{sec:deploy}). Second,
even with a fresh cache, an
action head that only sees semantics cannot be reactive: short-horizon
correction requires fresh, low-level, spatially grounded input at the same
rate as the control loop.

\method{} addresses both failures with three ingredients.
\textbf{(i)~Age-conditioned cache reuse.} The expert is conditioned on the
\emph{measured} age $\tau$ of the cache (the time since the last refresh)
alongside the flow time and the current proprioceptive state.
Training samples the cache age $\Delta$ from the physical episode history,
so the policy sees the full staleness distribution it will encounter at
deployment, including the contradiction between stale semantics and fresh
geometry that it must learn to arbitrate.
\textbf{(ii)~A fresh visual stream on the fast clock.} All fast-moving
visual information enters the expert as a pluggable token block---%
appended to the expert's input sequence and attended by its
self-attention---and is refreshed wholesale at every re-planning step:
motion-aware image tokens from a frozen encoder pretrained on robot video
(masked reconstruction, optical flow, and depth, so inter-frame motion and
geometry are learned exactly where the stream is consumed). Slower-reading
signals attach through gated cross-attention with zero-initialized
output projections; the cache age $\tau$ enters through adaptive
layer normalization (AdaRMS) and proprioception as a dedicated expert
input token.
\textbf{(iii)~A chain-of-thought-supervised anchor.}
The slow backbone carries a short chain-of-thought region appended after
the prompt; because attention is causal, the features the expert consumes
are unchanged by the appended region, so the reasoning shapes what the
anchor knows without entering the action path---and a verdict head built
on the same supervision gates execution under instruction--scene mismatch
(Sec.~\ref{sec:mismatch}).

We evaluate on LIBERO~\cite{libero}, zero-shot on the LIBERO-plus
perturbation benchmark~\cite{liberoplus}, on the LIBERO-PRO
perturbation benchmark~\cite{liberopro}, and on a four-platform
real-robot suite.
Our contributions are:
\begin{itemize}
\item An asymmetric-frequency VLA architecture that preserves the
layer-wise semantic coupling of shared-attention VLA backbones while
reusing the semantic cache across control steps.
\item Staleness-aware cache reuse: measured-age conditioning combined with
physical-clock staleness-distribution training and a deployment-homomorphic
corruption curriculum, evaluated as a single-variable training ladder on
the asynchronous deployment stack that recovers 61.4\%$\to$95.5\% under
slowed semantic refresh on LIBERO-Spatial (Table~\ref{tab:ladder}).
\item A unified injection interface for the fast loop: the visual
stream---motion-aware image tokens---enters the expert through
one pluggable token-block interface (sequence concatenation; gated
cross-attention with staggered schedules and zero-initialized gating for
additional streams), so streams can be added, upgraded,
or dropped without backbone surgery; plus a robot-video
pretraining recipe (masked reconstruction + optical flow + depth) that
makes the frozen fast-stream encoder motion- and geometry-aware.
\item A deployment validation of the decoupled stack under physical
asynchrony: \emph{soft-nudge}, a forward-only chunk-boundary correction
that matches
autograd-based real-time-chunking guidance without backward passes---making
per-control-step re-planning the best operating point (97.1\%)---together with a
conflict-geometry analysis locating what inference-time correction can and
cannot fix.
\item An instruction--scene consistency study: zero-cost negative episodes
built from cross-suite instruction re-assignment, a measured
\emph{achievability-prior} failure (the policy manipulates the scene under
mismatched instructions), and a chain-of-thought verdict gate that freezes
the policy exactly on mismatched instructions at zero cost to feasible
success (Sec.~\ref{sec:mismatch}).
\end{itemize}

% ============================================================================
\section{Related Work}
\label{sec:related}

\subsection{Vision-Language-Action Models}
VLA models inherit pretrained VLMs~\cite{qwen25vl,qwen3vl} and fine-tune them
for control, either by decoding actions as text~\cite{rt1,rt2}, by attaching
regression heads~\cite{openvla}, or by generative action
decoders~\cite{pi0,pi05,diffusionpolicy}. $\pi_0$~\cite{pi0} popularized the
architecture we build on: a VLM prefix (frozen in $\pi_0$) and a smaller flow-matching
action expert coupled through \emph{shared attention}, where every expert
layer attends the KV of the corresponding VLM layer. $\pi_{0.5}$~\cite{pi05}
extends this with open-world generalization. All of these systems re-run the
full VLM at every re-planning step; none conditions on the age of its
semantic context.

\subsection{Hierarchical and Dual-Rate Robot Policies}
Several systems decouple slow reasoning from fast control. Hi
Robot~\cite{hierobot} lets a VLM restructure user commands for a fast
low-level policy. Helix~\cite{helix} runs a 7B VLM at 7--9\,Hz and an 80M
visuomotor policy at 200\,Hz, communicating through a single fixed-size
latent vector. G0~\cite{g0} couples a VLM planner with a separately
trained VLA executor---two full models, each carrying its own
vision--language backbone, joined through subtask instructions.
RoboDual~\cite{robodual} and latent-bridge approaches~\cite{lcb}
distill a generalist's output into a specialist's conditioning;
Fast-in-Slow~\cite{fis} repurposes the VLM's final blocks as the fast
system, which consumes the slow system's latest latent regardless of
its age.
These works build the dual system by \emph{composing} models: two
separately parameterized, separately trained systems whose fast side sees
the slow one only through language or a fixed latent that bottlenecks what
crosses. \method{} instead \emph{splits} a single VLA into its two loops:
the slow and fast loops are the two halves of one shared-attention model,
coupled layer-wise through the full cached context, and the staleness the
split introduces is made \emph{explicit and learnable} through measured-age
conditioning rather than hidden behind a fixed latent. To our knowledge, no prior VLA work
conditions its action expert on the measured age of its semantic context or
studies success as a function of that age.

\subsection{Latency, Chunking, and Asynchrony}
Action chunking with temporal ensembling~\cite{act} smooths execution across
overlapping chunks; real-time chunking~\cite{rtc} mitigates inference latency
at deployment by inpainting actions that will execute while the next chunk is
computed. These are inference-time remedies orthogonal to ours: they assume a
policy queried at high rate, which is exactly what \method{}'s fast expert
enables.

\subsection{Structured Reasoning in VLA}
Embodied chain-of-thought~\cite{ecot} interleaves free-text reasoning with
control, placing explicit reasoning text on the inference critical path. We
instead co-train the chain-of-thought with the control objective: the
reasoning capability is internalized in the anchor during training, and
explicit CoT generation at deployment is optional. Our slow backbone
carries a short chain-of-thought region trained with a language-modeling
objective;
causal attention keeps the expert's
input features bitwise identical, so the reasoning shapes the anchor's
representation without entering the action path, and a verdict head built on
the same supervision gates execution under instruction--scene mismatch
(Sec.~\ref{sec:mismatch}).

\subsection{Self-Supervised Visual Pretraining for Control}
Robot-domain visual pretraining distills web-scale self-supervision into
control-friendly features, from stage-one objectives (R3M~\cite{r3m},
Voltron~\cite{voltron}) to masked autoencoding~\cite{mae}. Our fast-stream
encoder follows the masked-autoencoder recipe but adds explicit
\emph{inter-frame motion} and \emph{geometry} objectives---dense optical
flow against a frozen teacher~\cite{waft}, and per-pixel depth against a
frozen depth teacher~\cite{depthv2}---precisely because the stream exists to carry
short-horizon motion to the expert; the pretrained encoder is then consumed
frozen through cross-attention injection (Sec.~\ref{sec:pretrain}).

% ============================================================================
\section{Method}
\label{sec:method}

\subsection{Overview}
\begin{figure*}[t]
\centering
\includegraphics[width=\textwidth]{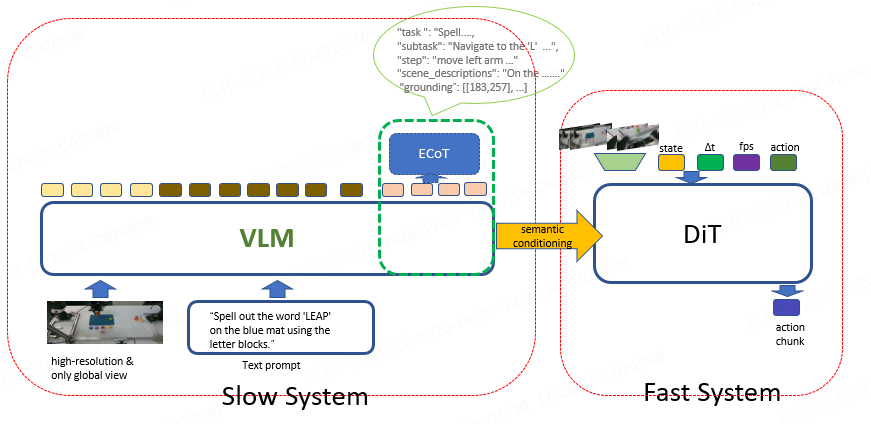}
\caption{\method{} overview. \emph{Slow system:} the VLM runs at
\tslow{} over a token sequence that appends, after the prompt and the
observation, the chain-of-thought region; it produces the
cached context $\mathcal{C}$ and the chain-of-thought-supervised anchor
features, and its reasoning region is emitted as the structured output
shown. Because attention is causal, the prompt-region features the expert
consumes are unchanged by the appended block (Sec.~\ref{sec:slow}).
\emph{Fast system:} the flow-matching expert runs at \tfast{}, consuming
$\mathcal{C}$ \emph{and} fresh fast-stream features through one
pluggable token-block interface: image tokens from the frozen
motion-aware encoder (dual-frame, robot-pretrained with reconstruction,
optical flow, and depth). The measured cache age
$\tau$ conditions the expert via AdaRMS, proprioception enters as a
dedicated input token, and the control rate $f$ is passed to the action
head; training
samples $\tau$ from the physical episode history.}
\label{fig:overview}
\end{figure*}

\method{} separates a VLA policy into a \emph{slow loop} at frequency
\tslow{} and a \emph{fast loop} at frequency \tfast{}, with
\tslow{}$\,\ll\,$\tfast{}. The slow loop runs a Qwen2.5-VL-2B~\cite{qwen25vl}
VLM $\Phi$---initialized from the pretrained checkpoint and, depending on
the task, kept frozen, adapted with LoRA, or fully fine-tuned---that
encodes the instruction, a camera observation, and its own short
autoregressive output, and produces the \emph{cached context} $\mathcal{C}$:
the features it has processed for the current scene prefix, held fixed until
the next refresh. The concrete form of $\mathcal{C}$ is exchangeable---%
per-layer key--value pairs read by shared attention, or backbone features
read by cross-attention---and nothing downstream depends on which. The fast
loop runs a flow-matching
action expert $E_\theta$ that, at every control step, denoises an action
chunk while attending to $\mathcal{C}$ and to freshly encoded fast-stream
features (motion-aware image tokens), all entering through one pluggable
token-block interface. The only quantity coupling the two loops besides
$\mathcal{C}$ is the measured cache age $\tau$.
Fig.~\ref{fig:overview} illustrates the architecture.

\subsection{Slow System: The Semantic Anchor}
\label{sec:slow}

\subsubsection{Prompt and cache}
At refresh time $t_r$, the VLM prefix contains the language instruction and
the current camera observation. Crucially, the robot state is \emph{not}
part of the prompt: embedding state as text in a cached prefix is
self-defeating, since the cached copy becomes wrong the moment the robot
moves. State enters the fast loop instead (Sec.~\ref{sec:fast}). The prefix
therefore stays semantically valid under cache reuse: scene layout and task
identity, not instantaneous proprioception.

\subsubsection{Chain-of-thought supervision}
The backbone is a causal language model, and its structured output is a
short \emph{chain-of-thought} (CoT) region appended after the prompt: a
fixed set of fields, generated autoregressively and supervised with a
language-modeling (cross-entropy) loss through an untied LM head. Because
attention is causal and positions are relative, the prompt-region features
that the action head consumes are bitwise identical with and without the
appended block: CoT supervision shapes what the anchor \emph{knows}---the
features it hands the expert---without adding anything to the action path,
and no CoT needs to be generated at deployment unless its output is
consumed. We use the region in two roles: as training-side feature
supervision, and---through a verdict head trained on the same
supervision---as a deployment-side gate under instruction--scene mismatch
(Sec.~\ref{sec:mismatch}).

\subsubsection{Cache refresh policy}
The cache is refreshed (i)~periodically every $T_{\mathrm{ref}}$ seconds,
(ii)~on trigger events---large frame difference or optical flow at the image
level, gripper open/close events, or subtask changes---and (iii)~on
episode reset. The reported deployments exercise (i) and (iii) on the
request-gated schedule of Sec.~\ref{sec:setup}; the trigger path is
supported by the design but not exercised in any reported cell. Refreshes
are executed asynchronously; the control loop never
blocks on the VLM. Each control step reads the \emph{measured} cache age
$\tau = t - t_{\mathrm{refresh}}$, which accounts for real scheduling jitter.

The fast-stream encoder (Sec.~\ref{sec:pretrain}) is pretrained offline
and frozen at deployment; the VLM starts from the pretrained checkpoint
and is kept frozen, adapted with LoRA, or fully fine-tuned as the task
requires. The slow loop's runtime jobs are the periodic refresh and, when
consumed, the structured output, all amortized over many control steps.

\subsection{Fast System: The Reactive Expert}
\label{sec:fast}

The expert follows the attention-based coupling of $\pi_0$~\cite{pi0}:
expert tokens attend the cached context $\mathcal{C}$ and each other, with
the concrete form of the coupling left open (per-layer shared attention as
in $\pi_0$, or cross-attention to the cached features). We turn the expert
into a \emph{fusion hub} with two mechanisms: a
scalar conditioning vector for everything scalar, and a set of
\emph{modalities}---one pluggable token-block interface through
which every stream-valued signal enters.

\subsubsection{Conditioning vector}
$\mathbf{c} = \mathrm{MLP}([\,t,\; \mathrm{enc}(\tau)\,])$, where
$t$ is the flow time and $\tau$ the measured cache age; $\mathbf{c}$
modulates every
expert layer via AdaRMS~\cite{dit,film}. The age is encoded as a
sinusoidal embedding of $g(\tau)=\log(1+\tau)$ seconds
(Sec.~\ref{sec:staleness}).
The proprioceptive state $s_t$ enters as a dedicated input token
alongside the action chunk, attended by the expert's self-attention.
Moving $s_t$ out of the prompt is what makes arbitrary cache
ages representable: the prefix no longer carries any instantaneous
quantity. The age encoding enters through its own
zero-initialized projection (Sec.~\ref{sec:staleness}), so staleness
conditioning, too, is a no-op at initialization. The fast loop's control
frequency $f$ is passed to the action head as an additional input.

\subsubsection{Modalities: one interface, many sources}
Each modality $m$ provides a fixed-length token block
$\mathbf{M}_m \in \mathbb{R}^{N_m \times d_m}$ (per camera for multi-view
streams), refreshed wholesale at every re-planning step, and enters the
expert through one of two attention forms---both pluggable attachments,
so a modality can be added, upgraded, or dropped without backbone surgery.
\emph{(a)~Sequence concatenation}: the block is projected to the expert
width, given learned positional embeddings, appended to the expert's
input sequence, and consumed by its self-attention at every block---%
today's fast visual stream, the highest-rate consumer.
\emph{(b)~Gated cross-attention}: the block is read through a dedicated
cross-attention whose output projection is \emph{zero-initialized}
(Flamingo-style gated injection~\cite{flamingo}), so attaching it is a
no-op at initialization and the pre-attachment behavior is recoverable;
a per-block stride schedule ($\ell \ge o_m$,
$(\ell - o_m) \bmod e_m = 0$~\cite{groot}) controls how often a modality
is consulted (the cached context: every second block, in the deployed
configuration). A stream is encoded \emph{once} per
re-planning cycle and its features are reused across all Euler steps,
so injection cost is linear in the number of attached modalities and
independent of the camera count at deployment batch size
(Sec.~\ref{sec:efficiency}). All temporal aggregation happens \emph{before}
tokenization: perception streams are replaced wholesale each cycle, and the
expert itself never aggregates time.

The fast visual stream instantiates form~(a) today; form~(b) admits
further batch-side streams:
\begin{itemize}
\item \textbf{Motion-aware image tokens} (today's visual stream): the
frozen encoder of Sec.~\ref{sec:pretrain} encodes a fresh dual-frame pair
per camera; each camera contributes $196$ tokens ($224^2$/patch-16), so
$V$ cameras give $V{\times}196$ feature tokens, projected to the expert
width with a patch positional embedding shared across cameras.
\end{itemize}

\subsubsection{Cost}
The expert suffix stays minimal---the flow-matching chunk ($H{=}20$ action
tokens plus a proprioceptive state token)---so the joint sequence length is
independent of the number of attached modalities; streams are paid for
only at their injection layers. The expert is a $1536$-wide, 32-block transformer; with
4 Euler steps it leaves ample headroom for control periods of
50\,ms and below (Sec.~\ref{sec:efficiency}).

\subsection{Motion-Aware Pretraining of the Fast Visual Stream}
\label{sec:pretrain}

The fast loop needs a visual encoder of its own, and three requirements
separate it from the VLM's vision tower: it must run every re-planning
cycle (compact, launch-cheap); its features must be \emph{spatially dense
and motion-informative} rather than language-aligned semantics (they are
consumed by the expert's cross-attention, not by a language model); and it
must see the robot's own visual domain. We therefore pretrain a compact
encoder on multi-source robot video, then freeze it into the image
modality of Sec.~\ref{sec:fast}.

\subsubsection{Architecture}
A $\sim$23M-parameter encoder: a convolutional stem with four downsampling
stages followed by three wide transformer layers with rotary position
embeddings, so the positional encoding is resolution-agnostic (the same
weights run at reduced resolution for ablations). Each sample is a
\emph{dual-frame pair} $(o_{t-\delta}, o_t)$ whose two frames are fused
into a 6-channel input at the stem (early fusion), yielding $196$ tokens
per camera at $224^2$/patch-16; the deployed geometry is identical to the
pretraining geometry, so no representation shift is introduced at
deployment.

\begin{figure}[t]
\centering
\includegraphics[width=\columnwidth]{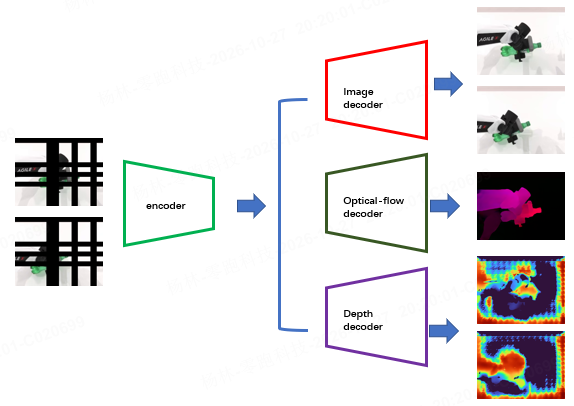}
\caption{Fast-stream pretraining objective. A dual-frame pair
$(o_{t-\delta}, o_t)$ is fused into a 6-channel input at the stem; a
${\sim}49\%$ row/column strip mask is applied jointly to both frames
(pair-level tube masking), and the visible patches go through the compact
encoder. Three lightweight decoders share the encoded features: the
reconstruction decoder fills in \emph{both} frames' masked patches, the
optical-flow decoder predicts the dense $o_{t-\delta}\!\to\!o_t$ motion of
masked patches, and the depth decoder predicts per-pixel depth for each
frame. Panels show a held-out RoboTwin sample; decoder outputs are the
training-time ground truth (WAFT flow, Depth Anything V2 depth, and the
raw frames).}
\label{fig:fastmae}
\end{figure}

\subsubsection{Objectives: reconstruction, motion, and depth}
Three masked-autoencoder-style~\cite{mae} objectives share the encoder
(Fig.~\ref{fig:fastmae}).
(i)~\emph{Masked reconstruction}: ${\sim}49\%$ of patches (a row/column
strip mask) are held out, and a lightweight decoder reconstructs
\emph{both} frames of the pair from the merged features.
(ii)~\emph{Inter-frame motion}: a flow decoder predicts, for every masked
patch, the dense optical flow $o_{t-\delta} \to o_t$ in pixels, supervised
by per-pixel end-point error against ground truth generated in the data
loader by a frozen off-the-shelf flow teacher (WAFT~\cite{waft}). Flow
supervision penetrates the same merged features the reconstruction uses,
so the encoder is explicitly taught \emph{what changed between the two
frames}---the quantity the reactive loop cares about.
(iii)~\emph{Depth}: a depth decoder predicts, for every masked patch, the
per-pixel depth of each frame of the pair, supervised by per-pixel depth
error against ground truth generated in the same data loader by a frozen
off-the-shelf depth teacher (Depth Anything V2~\cite{depthv2}).
Reconstruction teaches appearance, flow teaches \emph{what changed}, and
depth teaches \emph{how far}---the contact-relevant geometry that
appearance-only features drop, and that survives the lighting and texture
shifts the reactive loop must operate through. At deployment the
fresh dual-frame pair (gap $\delta = 0.2$\,s, matching pretraining) hands
the expert short-horizon motion for free.

\subsubsection{Data}
The encoder must serve every robot embodiment, so pretraining draws
from 62 robot datasets totalling $\sim$319k episodes and $\sim$73M frames:
nine curated multi-camera suites ($\sim$63.5k episodes, $\sim$45M frames,
2--3 cameras each) mixed with 53 public OXE sources ($\sim$256k episodes,
$\sim$28M frames), holding the physical pair gap $\delta$ fixed across
sources via fps-normalized frame strides. The motivation is empirical: an
identical architecture initialized from a driving-domain checkpoint is
nearly scene-blind on robot video (cross-scene feature cosine $0.92$),
whereas after robot-video pretraining the same probe drops to $0.58$ while
same-scene features stay stable ($0.99$)---robot-data pretraining, not
generic pretraining, is what makes the frozen encoder discriminative.

\subsection{Staleness-Aware Training}
\label{sec:staleness}

\subsubsection{Random cache-age sampling}
For each training sample at time index $i$ of an episode, we sample
\[
\Delta \sim \mathcal{U}\!\left[0,\; \Delta_{\max}\right]\ \text{(seconds)},\qquad
\Delta_{\max} = T_{\mathrm{ref}} + \epsilon_{\mathrm{jit}},
\]
retrieve the anchor frame at index $i - \lfloor \Delta \cdot \mathrm{fps}\rfloor$
\emph{from the same episode}, and clamp at the episode start, using the
\emph{effective} age $\Delta^{\mathrm{eff}}$ (not the sampled one) for
encoding. Sampling is performed on the fly at every epoch pass; $\Delta=0$
is included, matching deployments where a control step immediately follows a
refresh. The effective age is encoded as a sinusoidal embedding of
$g(\tau)=\log(1+\tau)$ and injected into the conditioning vector
$\mathbf{c}$ through a zero-initialized projection.

\subsubsection{Age-ramped pixel corruption}
Random ages define \emph{when} the snapshot is unreliable, not \emph{how}
staleness manifests. A deployed stale snapshot is not uniformly wrong: scene
layout tends to persist while arm geometry and object poses drift. We
therefore corrupt the stale snapshot itself: a contiguous rectangle of the
VLM image (before the vision tower) is erased to gray, with area ramping
continuously in the sampled age,
$\mathrm{frac} = \mathrm{frac}_{\max}\cdot
\mathrm{clamp}\!\big((\Delta - t_0)/(\Delta_{\max}-t_0)\big)$,
capped below the full image. The corruption is \emph{deployment-homomorphic}
(a stale snapshot is a partially invalid image, not a missing one), and it is
applied at the pixel level so every downstream consumer---vision tower,
cached prefix features, text grounding---sees the corrupted image. Below $t_0$ the input
is bit-identical, leaving the $\tau{=}0$ operating point unperturbed. The
motivation is empirical: trained only on uncorrupted stale frames, the model
hedges---it averages its way through contradictions between stale semantics
and fresh geometry instead of arbitrating them.

\subsubsection{What staleness training teaches}
Random ages turn the policy into an explicit function
$\pi(a \mid o_{\mathrm{fast}}(t),\, \mathcal{C}(t-\tau),\, \tau)$ of all
three inputs. Three behaviors are thereby trained rather than hoped for:
(i)~\emph{arbitration}: when cached semantics and fresh geometry disagree
(they do, by construction, for large $\tau$), the expert learns to weight the
fresh stream; (ii)~\emph{age-appropriate trust}: the $g(\tau)$ encoding lets
the expert modulate how far it plans against possibly outdated semantics;
(iii)~\emph{distribution coverage}: deployment jitter and refresh-thread
latency land inside the trained age range by construction of
$\Delta_{\max}$.

\subsection{Training and Inference}
\label{sec:training}

Training uses the standard flow-matching objective
of~\cite{pi0,lipman} on the
truncated action dimensions,
$\mathcal{L} = \mathbb{E}\,\|u_t - v_\theta(x_t,\cdot)\|^2$ over the real
action dimensions only, with the chain-of-thought language-modeling loss
added when CoT annotations are available (Sec.~\ref{sec:slow}). Because the fast-stream encoder is frozen, its features can be
precomputed; when the VLM is likewise kept frozen, so can the cached
prefix features. This both speeds training and permits full-precision
expert optimization without the memory footprint of the backbones.
At inference, each
environment holds its own cache; a re-planning cycle runs the prefix,
encodes the fresh fast stream once (its features are reused across all
Euler steps), then denoises with the measured $\tau$; (optionally)
overlapping chunks are fused by temporal voting~\cite{act,rtc}.

% ============================================================================
\section{Experiments}
\label{sec:exp}

\subsection{Setup}
\label{sec:setup}
\textbf{Benchmarks.} LIBERO~\cite{libero} (goal, spatial, object, long)
built on robosuite~\cite{robosuite}; LIBERO-plus~\cite{liberoplus}, which
extends the same four suites with perturbed task variants over seven
dimensions (Sec.~\ref{sec:plus}); LIBERO-PRO~\cite{liberopro}, which
perturbs the same four suites along object, position, language, task, and
environment dimensions with an official baseline protocol
(Sec.~\ref{sec:pro}); and a real-robot suite on four
platforms---two dual-arm workstations (Realman, Galaxea), a Dobot Atom
mobile dual-arm humanoid, and an in-house humanoid upper body
(Sec.~\ref{sec:real}); seven tasks, 50 trials per task.
\textbf{Baselines.} (1)~frequency-matched fine-tuning with same-frequency
joint inference (full backbone re-encoding
at every step; the compute-unconstrained anchor and the reference row of
Sec.~\ref{sec:deploy}); (2)~the same policy evaluated under slowed semantic
refresh without staleness-aware training (the naive-reuse regime); (3)~real-time
chunking~\cite{rtc} stacked on \method{} (Sec.~\ref{sec:smooth}).
\textbf{Protocol.} 50 episodes per task with fixed initial states (single
seed); the deployment cells of Secs.~\ref{sec:deploy}--\ref{sec:smooth}
score all ten spatial tasks (500 episodes per cell). Re-planning rate held fixed across variants when comparing
refresh rates, and vice versa; the measured cache-age distribution is
reported per regime. The deployment-line sections
below additionally run a chunk-level asynchronous loop with leftover-prefix
passing, detailed in Appendix~\ref{app:protocol}. Unless stated otherwise,
the LIBERO models in all tables are trained without chain-of-thought
supervision; CoT supervision and the verdict gate are studied separately
in Sec.~\ref{sec:mismatch}.
\textbf{Deployed-age protocol (measured).} All LIBERO numbers below come
from the asynchronous deployment stack. Because the contribution is a
\emph{timescale decoupling}, we parameterize operating points by the
semantic-condition refresh period $T_{\mathrm{slow}}\in\{0.2,0.5,1,2\}$\,s
against a fixed re-plan clock $T_{\mathrm{replan}}{=}0.5$\,s ($R{=}10$ of
each chunk's 20 actions are consumed at the 50\,ms control period)---under
request-gated stale-clock refresh on observation timestamps: a request due
for refresh takes a fresh context ($\tau\approx0$), intermediate requests
reuse the cache, so the measured context age cycles
$0{\to}0.5{\to}1.0{\to}1.5$\,s under the 2\,s refresh (mean ${\approx}0.75$\,s,
p99 $\le 1.5$\,s, inside the trained range $\Delta_{\max}{=}2$\,s), and
every re-planning request reports the \emph{measured} age of the context it
used. The fast stream keeps its
data-anchored pair gap ($\delta{=}0.2$\,s, Sec.~\ref{sec:pretrain}). All
cells share the same
denoiser and gating.

\textbf{Synchronous baselines under decoupled clocks.} Synchronous systems
are released as such, and our harness reproduces exactly that
regime at the matched operating point (DiT:VLM $1{:}1$): a synchronous
model's matched cell sits at its published
operating point (no handicap), and the decoupled fast loop we run for it reuses
the cached backbone output while re-encoding proprioception at every
control step---at the matched operating point this path is bitwise identical to its native
synchronous forward pass. The slower-refresh cells (DiT:VLM $2{:}1$ and
$4{:}1$) therefore measure a
\emph{structural} property, not a harness artifact: a single-pathway
system whose control path runs through the VLM output has no fresh visual
input at control rate once the semantic loop slows ($\pi_{0.5}$ embeds
the robot state as prompt text, so its cache freezes proprioception as
well; GR00T's head re-encodes state but sees no images between
refreshes), whereas \method{}'s fast stream keeps per-step visual
grounding by construction. Staleness-aware training is part
of our method claim rather than a hidden advantage; whether synchronous
baselines would recover under staleness curricula of their own is an open
question we do not claim to answer.

\subsection{LIBERO Four-Suite Comparison}
\label{sec:anchor}

\begin{table*}[t]
\centering
\caption{Published LIBERO four-suite success (\%), transcribed from each
method's original publication (as aggregated in the public X-VLA benchmark
table). Published rows follow each method's own evaluation protocol and are
not commensurable with one another; the table is public context, not a
controlled comparison. The Leap-VLA row reports our fine-tuned model,
measured on the asynchronous deployment stack used throughout this paper
at the matched-rate operating point (DiT:VLM scheduling ratio $1{:}1$;
$50$ episodes per task; Appendix~\ref{app:protocol}).}
\label{tab:libero_pub}
\begin{tabular}{lccccc}
\toprule
Model & Goal & Spatial & Object & Long & Avg \\
\midrule
$\pi_0$~\cite{pi0} & 94.4 & 98.0 & 96.8 & 88.4 & 94.4 \\
InternVLA-M1~\cite{internvlam1} & 93.8 & 98.0 & 99.0 & 92.6 & 95.9 \\
\textbf{Leap-VLA (ours)} & 97.2 & 97.5 & 98.0 & 92.4 & 96.3 \\
$\pi_{0.5}$~\cite{pi05} & 98.0 & 98.8 & 98.2 & 92.4 & 96.9 \\
GR00T-N1.7~\cite{groot} & 97.5 & 97.7 & 98.5 & 94.4 & 97.0 \\
OpenVLA-OFT~\cite{openvlaoft} & 97.9 & 97.6 & 98.4 & 94.5 & 97.1 \\
Fast-WAM~\cite{fastwam} & 97.0 & 98.2 & 100.0 & 95.2 & 97.6 \\
Motus~\cite{motus} & 96.6 & 96.8 & 99.8 & 97.6 & 97.7 \\
Being-H0.5~\cite{beingh05} & 99.0 & 99.2 & 98.2 & 96.2 & 98.2 \\
EO1~\cite{eo1} & 99.2 & 99.7 & 99.8 & 94.8 & 98.4 \\
Cosmos Policy~\cite{cosmospolicy} & 98.2 & 98.1 & 100.0 & 97.6 & 98.5 \\
LingBot-VA~\cite{lingbotva} & 97.2 & 98.5 & 99.6 & 98.5 & 98.5 \\
Xiaomi-Robotics-0~\cite{xiaomirobotics0} & 98.8 & 98.8 & 100.0 & 97.2 & 98.7 \\
G0.5~\cite{g05} & 98.6 & 98.4 & 100.0 & 98.6 & 98.9 \\
\bottomrule
\end{tabular}
\end{table*}

Table~\ref{tab:libero_pub} is the anchor the rest of this section departs
from, with one caveat governing the reading: each published row reports
its own evaluation stack---episode budgets, initial-state distributions,
and success checks differ across papers---so the table is public context,
not a controlled comparison. Within that caveat, the published four-suite
averages span $94.4$ ($\pi_0$~\cite{pi0}) to $98.9$ (G0.5~\cite{g05}),
with Xiaomi-Robotics-0~\cite{xiaomirobotics0} ($98.7$) close behind. Our
fine-tuned model posts $96.3$---above $\pi_0$ ($94.4$), $0.7$ points
behind GR00T-N1.7~\cite{groot} ($97.0$), and $2.6$ behind the published
top---measured, unlike any published row, through the full asynchronous
deployment stack at its matched-rate operating point. Finally, the Long
column is where published numbers spread widest ($88.4$ to $98.6$)---the
clearest sign that protocol choices, not only model quality, move the
hardest suite. This frontier is the level the decoupled system must reach
without paying for its structure.

\begin{table}[t]
\centering
\caption{Leap-VLA on all four LIBERO suites as the semantic loop slows
($50$ episodes per task, fixed seed; the client re-plans every
$0.5$\,s and the semantic condition is refreshed every 0.5/1/2\,s,
i.e., on every 1st/2nd/4th re-plan; stale-clock gating on observation
timestamps, Appendix~\ref{app:protocol}).}
\label{tab:libero_refresh}
\begin{tabular}{lccc}
\toprule
 & \multicolumn{3}{c}{Semantic-condition refresh period $T_{\mathrm{slow}}$} \\
\cmidrule(lr){2-4}
Suite & 0.5\,s & 1\,s & 2\,s \\
 & \emph{DiT:VLM $1{:}1$} & \emph{$2{:}1$} & \emph{$4{:}1$} \\
\midrule
Goal    & 97.2 & 95.8 & 95.6 \\
Spatial & 97.5 & 95.9 & 95.5 \\
Object  & 98.0 & 98.0 & 96.2 \\
Long    & 92.4 & 91.2 & 90.8 \\
\midrule
Average & 96.3 & 95.2 & 94.5 \\
\bottomrule
\end{tabular}
\end{table}

Holding the deployment stack fixed and slowing only the semantic loop
(Table~\ref{tab:libero_refresh}), lengthening the semantic-condition
refresh period from 0.5\,s to 1\,s---halving the slow loop's
compute---costs $1.1$ points on average; doubling it
again to 2\,s costs a further $0.7$, and the degradation is uniform,
$1.6$--$2.0$ points per suite from matched to 2\,s (Long
$92.4\to90.8$, Object $98.0\to96.2$). The sections that follow ask the complementary
training-side question: which ingredients make a policy indifferent to
this slowdown.

\subsection{Training for the Deployed Clock}
\label{sec:deploy}

This section asks the deployment-facing question directly: on the real
asynchronous stack, which
\emph{training-side} ingredients recover the success that semantic
staleness destroys? We run this ablation on LIBERO-Spatial: the
configuration measured throughout is a vision--language backbone and a
flow-matching action expert, fine-tuned from the shared
four-suite LIBERO checkpoint,
and every cell is evaluated
through the full asynchronous deployment loop---a 50\,ms client clock,
re-planning every 0.5\,s, semantic prefix refreshed at the stated period,
stale-clock gating on observation timestamps. Each cell scores 500
episodes. Rows of Table~\ref{tab:ladder} add one ingredient at a time.

\begin{table*}[t]
\centering
\caption{Success (\%) on LIBERO-Spatial versus semantic-condition refresh
period (500 scored episodes per cell). Each row adds
one training ingredient to the row above. The frequency-matched reference
is fine-tuned at the matched period and never sees asynchronous data.
Re-planning is held at $T_{\mathrm{replan}}{=}0.5$\,s throughout; the
matched column evaluates at the
fine-tuning refresh period ($0.2$\,s).}
\label{tab:ladder}
\begin{tabular}{lccc}
\toprule
Training variant & \multicolumn{3}{c}{Semantic-condition refresh period $T_{\mathrm{slow}}$} \\
\cmidrule(lr){2-4}
 & 0.2\,s (matched) & 1\,s & 2\,s \\
\midrule
Frequency-matched fine-tuning (reference) & \textbf{97.5} & 82.9 & 61.4 \\
\quad $+$ asynchronous $\Delta t$ exposure & -- & 92.5 & 91.8 \\
\quad\quad $+$ measured-$\Delta t$ conditioning & -- & 94.5 & 92.9 \\
\quad\quad\quad $+$ stale-frame masking curriculum & -- & \textbf{95.9} & 95.5 \\
\bottomrule
\end{tabular}
\end{table*}

\textbf{(1) Exposure, not conditioning, is the main mechanism.} Training on
asynchronously timestamped data alone---with the conditioning inputs
disabled and nothing else changed---lifts the 2\,s cell from 61.4 to
91.8, recovering most of the 36.1-point gap: most of the deployed collapse
is a \emph{distribution} mismatch, and covering the deployed staleness
distribution during training closes most of it. \textbf{(2)~Conditioning sharpens the middle of the
range.} At a 1\,s refresh, where the reference still loses 14.6 points
(82.9 vs.\ 97.5 at its matched rate), measured-age conditioning recovers
11.6 of them (94.5) and the stale-frame masking curriculum (the
age-ramped corruption of Sec.~\ref{sec:staleness}) adds the best
asynchronous cell of the table: 95.9 at a 1\,s refresh, within 1.6 points
of the frequency-matched reference at its own rate (97.5)---asynchronous
training pays almost no synchronous tax at moderate rates.

\begin{table*}[t]
\centering
\caption{Success (\%) versus semantic-condition refresh period on the
deployment line
(LIBERO-Spatial, 500 episodes per cell---the same measured cells as
Table~\ref{tab:ladder}).
Asynchronous training (ours) is the full asynchronous training recipe of
that table; the range row spans the three
asynchronous training variants. Measured context
age follows the deployed-age protocol of Sec.~\ref{sec:setup}.}
\label{tab:ratio}
\begin{tabular}{lccc}
\toprule
 & \multicolumn{2}{c}{Semantic-condition refresh period} & \\
\cmidrule(lr){2-3}
 & 1\,s & 2\,s & $\Delta$(2\,s$-$1\,s) \\
\midrule
Measured context age (s) & $0/0.5$ alternating & $0{\to}1.5$ cycle & -- \\
Frequency-matched fine-tuning (reference) & 82.9 & 61.4 & $-21.5$ \\
Asynchronous training (ours) & 95.9 & 95.5 & $-0.4$ \\
\quad per-variant range & $[92.5,\,95.9]$ & $[91.8,\,95.5]$ & $[-1.6,\,-0.4]$ \\
\bottomrule
\end{tabular}
\end{table*}

Table~\ref{tab:ratio} reads Table~\ref{tab:ladder} along the ratio axis
alone. Doubling the refresh period from $1$\,s to $2$\,s costs the
frequency-matched reference $21.5$ points; the asynchronous family drops
$0.9$ on average, and no variant loses more than $1.6$ points. At 500
episodes per cell (binomial SE ${\approx}1.1$ points) the honest reading
is success-flat, not improvement: once the training
distribution covers the deployed staleness, doubling the refresh
period---i.e.,
halving the semantic loop's compute---is free. The per-variant cells and
the matched-rate anchor live in Table~\ref{tab:ladder}; the aggregate
answers the deployment question directly: the recipe, once fixed, does
not degrade again as the slow loop slows further.

\begin{table*}[t]
\centering
\caption{Success (\%) on LIBERO-Spatial at the
$T_{\mathrm{slow}}{=}2$\,s semantic-refresh operating point of
Table~\ref{tab:ladder}, varying the re-planning period
$T_{\mathrm{replan}}$---the clock at which a fresh action chunk is
requested, distinct from the semantic refresh (the $R{=}1$ columns re-plan
on every 50\,ms control step and execute a single action per re-plan).
Soft-nudge: $\eta{=}1$, exponential prefix schedule spanning the first
$10$ entries of the fresh chunk (Sec.~\ref{sec:smooth}).}
\label{tab:deploy}
\begin{tabular}{lccc}
\toprule
 & \multicolumn{3}{c}{Re-planning period $T_{\mathrm{replan}}$} \\
\cmidrule(lr){2-4}
Training variant & 0.5\,s ($R{=}10$; default) & 0.05\,s ($R{=}1$) & 0.05\,s ($R{=}1$) $+$ soft-nudge \\
\midrule
Frequency-matched fine-tuning (reference) & 61.4 & 26.9 & 39.2 \\
Asynchronous training ($+$ stale-frame masking) & 95.5 & 91.4 & \textbf{97.1} \\
\bottomrule
\end{tabular}
\end{table*}

Table~\ref{tab:deploy} composes the training recipe with the boundary
correction of Sec.~\ref{sec:smooth}. Re-planning every control step is the
reactive ideal---every executed action then conditions on the current
state---but for a chunked policy it replaces each executed trajectory with
a sequence of mutually independent chunk heads, and plan jitter dominates:
the reference loses 34.5 points (61.4$\to$26.9) and the asynchronous recipe
4.1 (95.5$\to$91.4). Soft-nudge supplies the missing cross-chunk
consistency---at a one-step horizon, leftover alignment is exactly temporal
ensembling of overlapping chunks~\cite{act}, computed forward-only---and
the asynchronous policy holds its default-schedule success at the most
reactive setting: 97.1\% at 0.05\,s vs.\ 95.5 at its
own 0.5\,s default, while the reference recovers
only a third of its loss (39.2). Faster re-planning becomes free only when
the training distribution and the boundary correction are both in place;
neither substitutes for the other.

\subsection{Chunk-Boundary Continuity and Forward-Only Correction}
\label{sec:smooth}

Success rates hide a failure mode of chunked execution: when a new chunk
replaces the one being executed, its first executed action can jump. With
0.5\,s re-planning these boundaries arrive twice a second, and on the
uncorrected asynchronous stack the boundary displacement averages
$1.9\times$ the mean intra-chunk displacement
(Table~\ref{tab:smooth})---a visible jerk exactly when the semantic loop
refreshes.

Real-time chunking~\cite{rtc} addresses this at inference time by
conditioning the new chunk on the \emph{leftover} prefix of the executing
one, either by inpainting the prefix region or by guidance, which adds, at
every denoising step, a correction proportional to the gradient of the
prefix-consistency error. Both variants reach the velocity field through
autograd: the correction backpropagates through the expert at every step,
which adds a backward pass to every denoise cycle and blocks export to
compiled runtimes. We therefore use a forward-only reformulation, \emph{soft-nudge}.
Let $\hat{x}_1 = x_t + (1-t)\,v_\theta(x_t, \cdot)$ be the one-step
endpoint prediction at flow time $t$, $\ell$ the leftover prefix, and $w$
the same three-segment prefix weights as RTC (1 on the frozen region,
exponentially decaying over the correction horizon, 0 beyond). Each Euler
step becomes
\begin{equation}
x_{t+\delta} \;=\; \hat{x}_1 \;+\; \eta\, w \odot (\ell - \hat{x}_1),
\label{eq:nudge}
\end{equation}
i.e., the step lands on its predicted endpoint and then blends a fraction
$\eta$ toward the leftover ($\eta{=}1$ throughout). This is the RTC
guidance update with the Jacobian transpose replaced by the identity, which
saves the backward pass; Appendix~\ref{app:conflict} measures the two
head-to-head and finds them behaviorally equivalent ($2.6\%$ vs.\ $2.4\%$
bridging on adversarial leftovers, equal suffix error), as expected when
the leftover is the policy's own recent output. Soft-nudge is one fused
multiply-add per step and action dimension: no backward pass, no extra
forward pass, no added latency, and the
computation graph is unchanged for TensorRT/NPU export.

\begin{table*}[t]
\centering
\caption{Chunk-boundary smoothness on the asynchronous stack, base
checkpoint: the 200-trajectory LIBERO evaluation.
Boundary metric: 6-DoF displacement at the first executed step of each new
chunk, normalized by the mean intra-chunk displacement; peak boundary =
per-episode maximum; ${\pm}$ is the 95\% CI over episodes.}
\label{tab:smooth}
\begin{tabular}{lcccc}
\toprule
Execution & boundary/intra $\downarrow$ & peak bnd. $\downarrow$ & RMS $|\Delta a|$ $\downarrow$ & path len. $\downarrow$ \\
\midrule
No correction (baseline) & 1.907$\pm$0.731 & 0.281$\pm$0.096 & 0.091$\pm$0.021 & 7.29$\pm$1.30 \\
RTC guidance (autograd)~\cite{rtc} & 1.194$\pm$0.271 & 0.207$\pm$0.076 & 0.086$\pm$0.023 & 6.96$\pm$1.33 \\
\textbf{Soft-nudge (ours, forward-only)} & \textbf{1.035$\pm$0.186} & \textbf{0.181$\pm$0.068} & \textbf{0.085$\pm$0.024} & \textbf{6.91$\pm$1.32} \\
\bottomrule
\end{tabular}
\end{table*}

\begin{figure}[t]
\centering
\includegraphics[width=\linewidth]{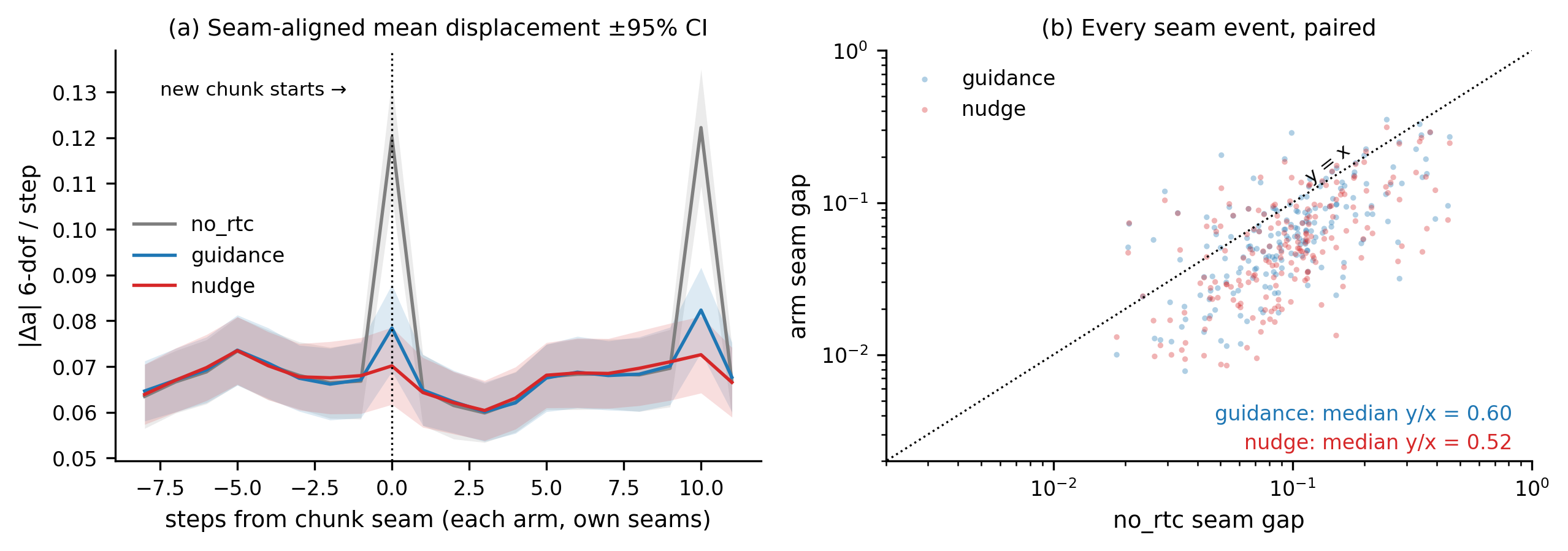}
\caption{Chunk-boundary analysis over the seam events of the 200-trajectory
evaluation behind Table~\ref{tab:smooth}.
(a)~Displacement at the boundary step, aligned over
all boundary events (mean $\pm$ 95\% CI; maxima 0.120 / 0.078 / 0.070; the
double spike reflects the 10-step chunk cadence). (b)~Boundary events
paired by ordinal position, corrected vs.\ uncorrected displacement
(log--log); the median corrected/uncorrected ratio is $0.60$ for RTC
guidance and $0.52$ for soft-nudge.}
\label{fig:seam-aligned}
\end{figure}

Table~\ref{tab:smooth} and Fig.~\ref{fig:seam-aligned} compare three
executions of the same policy on the 200-trajectory LIBERO evaluation:
no correction, RTC guidance, soft-nudge. Soft-nudge reduces the
boundary/intra ratio from 1.91 to 1.04
---the 95\% CI includes 1.0, i.e., a boundary step becomes statistically
indistinguishable from a mid-chunk step---and the peak boundary
displacement from 0.281 to 0.181, while overall action RMS and path length
barely move: the correction bends trajectories at boundaries, it does not
damp the motion. RTC guidance lands between the two
(1.19 / 0.207); per episode, soft-nudge attains the lowest peak boundary
displacement in the large majority of episodes and matches or beats
guidance on most episodes.
Appendix~\ref{app:crossckpt} reproduces the boundary
reduction on three further checkpoint families, and
Appendix~\ref{app:conflict} shows what the same corrections cannot do:
force execution to follow a leftover that genuinely conflicts with the
fresh observation.

\subsection{Efficiency and Scaling}
\label{sec:efficiency}

\textbf{Throughput accounting.} On shared hardware the two loops contend
for one GPU (serialized by a device lock), so achievable scheduling rates
follow from the per-invocation costs $t_{\mathrm{vlm}}$ and
$t_{\mathrm{act}}$: the same-frequency ceiling is
$1/(t_{\mathrm{vlm}}{+}t_{\mathrm{act}})$, and decoupling lifts the fast
loop to
$f_{\mathrm{fast}} \le \bigl(1 - f_{\mathrm{slow}}\,t_{\mathrm{vlm}}\bigr)/t_{\mathrm{act}}$,
where $f_{\mathrm{slow}}\,t_{\mathrm{vlm}}$ is the slow loop's duty cycle.
Three consequences we report rather than assume: (i)~running the slow loop
at saturation ($f_{\mathrm{slow}}{=}1/t_{\mathrm{vlm}}$) collapses the
bound back to the synchronous ceiling---decoupling pays only when the
slow loop runs below saturation (which the deployment study justifies) or on
dedicated hardware; (ii)~a request that lands on a due refresh waits
$t_{\mathrm{vlm}}{+}t_{\mathrm{act}}$, so fast-loop latency is reported
as p50/p99 under contention, not as the isolated $t_{\mathrm{act}}$;
(iii)~the frequency bought by slowing the slow loop is only useful if the
fast loop still observes the world: \method{}'s fast stream re-encodes
fresh images at every control step, so lowering $f_{\mathrm{slow}}$ trades
stale semantics for control frequency at a graceful success cost, whereas
a single-pathway system making the same trade loses all visual feedback at
control rate (Sec.~\ref{sec:setup}).
Table~\ref{tab:efficiency} instantiates this accounting on a single RTX
4090D (batch 1, deployment stack): measured per-request costs, the
same-frequency ceiling versus the decoupled bound, and p50 action latency
under contention; success as a function of the scheduling rates is
measured separately in
Secs.~\ref{sec:deploy}--\ref{sec:smooth}.

\begin{table*}[t]
\centering
\caption{Deployment cost on a single RTX 4090D (batch 1, the stack of
Sec.~\ref{sec:smooth}: bf16 semantic backbone, fp32 fast loop and action
expert; p50 over $\ge$30 requests after warm-up): the same
model and weights, only the schedule changes, each system at its best
measured configuration (denoise loop graph-captured on both sides).
Occupancy is the GPU duty cycle while serving the most reactive
re-planning clock of Table~\ref{tab:deploy} (one request per 50\,ms
control step, at that column's 2\,s semantic refresh).}
\label{tab:efficiency}
\setlength{\tabcolsep}{5pt}
\begin{tabular}{lccc}
\toprule
 & Single system (same-frequency) & \method{} (decoupled) & Gain \\
\midrule
Action latency per request & 73.9\,ms & \textbf{26.7}\,ms & $2.8\times$ \\
Compute per request (semantics amortized 1:4) & 73.9\,ms & \textbf{38.5}\,ms & $1.9\times$ \\
Max single-stream re-planning rate & 13.5\,Hz & \textbf{26}\,Hz & $1.9\times$ \\
GPU occupancy at the 0.05\,s clock & 148\,\% (infeasible) & \textbf{56}\,\% & feasible only decoupled \\
\bottomrule
\end{tabular}
\end{table*}

\textbf{The gain of decoupling, on the deployment stack.} Decoupling
changes neither the model nor the weights---only where the semantic
encoding runs---and the effect lands directly on the serving path. Waiting
for one action request costs the same-frequency system 73.9\,ms (semantic
refresh plus fast loop, serialized on the critical path) and the decoupled
system 26.7\,ms: $2.8\times$ per request. Counting the semantic cost at
its amortized share (11.8\,ms per fast request at the 1:4 ratio),
per-request compute drops from 73.9 to 38.5\,ms ($1.9\times$), and the
re-planning clock a single stream can serve extends from 13.5 to 26\,Hz.
Semantics stop being a per-action prerequisite and become a background
expense amortized at $1/f_{\mathrm{slow}}$.

\textbf{Feasibility, not speed.} The advantage concentrates at the
reactive end. Serving the most reactive column of Table~\ref{tab:deploy}---a
fresh plan every 50\,ms control step, each plan executing a single action
under that column's 2\,s semantic refresh---the
same-frequency joint occupies the GPU at 148\%: past 100\%, the control
loop cannot close in real time, so this clock is undeployable for the
same-frequency layout on hardware, whatever its simulated success (26.9).
The decoupled schedule occupies 56\%: the only configuration that serves
the reactive column in real time, and precisely the column where the
asynchronous policy holds 91.4 (97.1 with soft-nudge).

\textbf{Fast-stream scaling.} The fast stream is not the bottleneck:
encoding $V$ wrist-camera frame pairs costs 1.3/1.8/2.3\,ms at
$V{=}1/2/3$---the first camera costs 1.3\,ms, each additional one
0.5\,ms---so the fast-loop budget belongs almost entirely to the denoise
cycle, and the boundary-correction variants of Table~\ref{tab:smooth}
measure at parity with the uncorrected path.

\subsection{Zero-Shot Robustness on LIBERO-plus}
\label{sec:plus}

\begin{figure}[t]
\centering
\includegraphics[width=\linewidth]{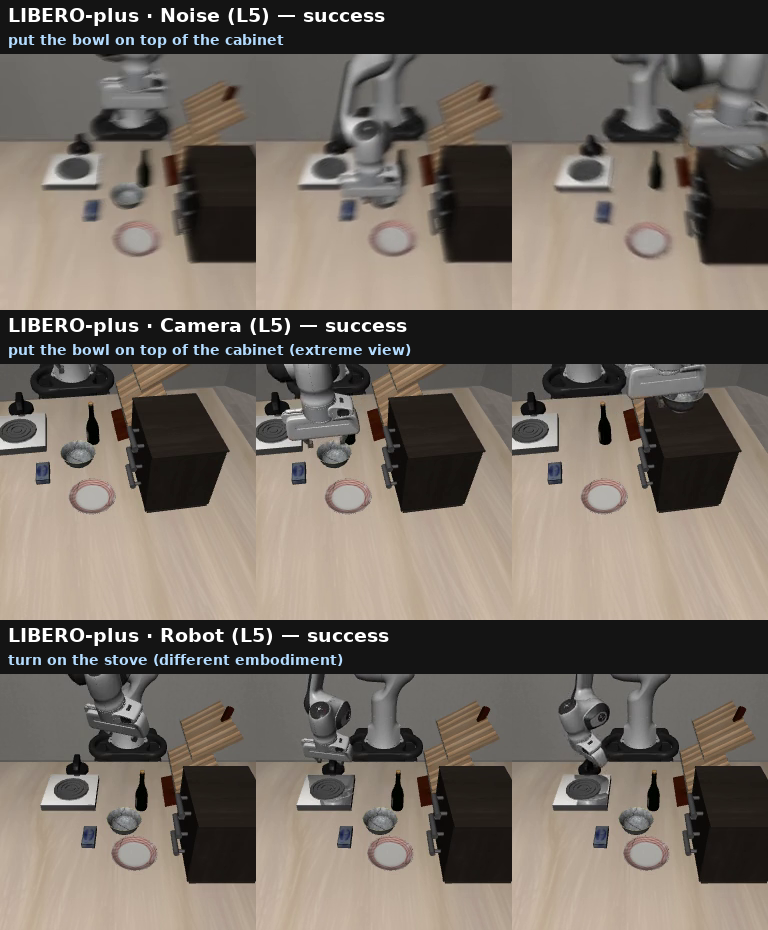}
\caption{Robustness to environment-level perturbations on LIBERO-plus
(difficulty L5), one successful episode per row. Top: heavy visual noise
corrupts the observations, yet the behavior is unaffected. Middle: an
extreme camera viewpoint. Bottom: an unseen robot embodiment. Under the
shared evaluation protocol, our model ranks 1st on Robot (68.4\%),
Camera (77.2\%), and Noise (82.6\%) among models trained without
LIBERO-plus data (69.2\% overall, first tier).}
\label{fig:plus-strengths}
\end{figure}

We submit the deployed system to LIBERO-plus~\cite{liberoplus}, which
extends the four LIBERO suites to 10{,}030 perturbed task variants across
seven dimensions---camera view, robot embodiment, language, lighting,
background, camera noise, object layout---graded over five difficulty
levels, one trial per task. The evaluated checkpoint is the shared four-suite
LIBERO model (600k steps), trained only on the
original LIBERO suites, so every perturbation dimension is zero-shot; it is
served by the same asynchronous deployment server as the deployment studies
(re-planning every 0.5\,s, no boundary correction). Table~\ref{tab:plus}
places the result on the benchmark's leaderboard. At 69.2\% total, the
model lands in the first tier of the group trained without LIBERO-plus
data (OpenVLA-OFT 69.6, ahead of $\pi_0$-Fast 61.6); it
leads the robot-embodiment dimension outright (68.4\%) and leads
camera view (77.2\%) and camera noise (82.6\%) within that group; per
suite it scores 81.0\% (object), 68.9\% (spatial), 68.8\% (goal), and
58.0\% (LIBERO-10), degrading from 80.4\% at difficulty level~1 to 53.3\%
at level~5. The dimension profile is complementary to the baselines':
lighting (53.5\%) and background (55.6\%) trail the strongest baselines by
35--40 points---the signature of a policy whose visual grounding is
geometric and motion-based but overfit to the photometric statistics of
its training distribution---exactly the dual of the baselines' weakness
in camera view and embodiment. Figure~\ref{fig:plus-strengths} shows one
successful level-5 episode on each of the three strongest dimensions;
Appendix~\ref{app:qual} illustrates the characteristic failures of the
two weakest.

\begin{table*}[t]
\centering
\caption{LIBERO-plus~\cite{liberoplus} leaderboard: success (\%) over the
full 10{,}030 perturbed tasks, grouped by perturbation dimension. All rows
except \method{} are quoted from the benchmark's public leaderboard under
the same protocol. Bold: best among models trained without LIBERO-plus
data; $\dagger$ marks the single entry fine-tuned on LIBERO-plus data
(not comparable). \method{} is trained only on the original LIBERO suites
and evaluated zero-shot on every perturbation.}
\label{tab:plus}
\begin{tabular}{lcccccccc}
\toprule
Model & Camera & Robot & Language & Light & Background & Noise & Layout & Total \\
\midrule
OpenVLA~\cite{openvla} & 0.8 & 3.5 & 23.0 & 8.1 & 34.8 & 15.2 & 28.5 & 15.6 \\
UniVLA~\cite{univla} & 1.8 & 46.2 & 69.6 & 69.0 & 81.0 & 21.2 & 31.9 & 43.9 \\
$\pi_0$~\cite{pi0} & 13.8 & 6.0 & 58.8 & 85.0 & 81.4 & 79.0 & 68.9 & 53.6 \\
OpenVLA-OFT$_w$~\cite{openvlaoft} & 10.4 & 38.7 & 70.5 & 76.8 & \textbf{93.6} & 49.9 & 69.9 & 55.8 \\
$\pi_0$-Fast~\cite{pi0fast} & 65.1 & 21.6 & 61.0 & 73.2 & 73.2 & 74.4 & 68.8 & 61.6 \\
OpenVLA-OFT$_m$~\cite{openvlaoft} & 55.6 & 21.7 & \textbf{81.0} & \textbf{92.7} & 91.0 & 78.6 & 68.7 & 67.9 \\
\method{} (ours) & \textbf{77.2} & \textbf{68.4} & 70.2 & 53.5 & 55.6 & \textbf{82.6} & 67.7 & 69.2 \\
OpenVLA-OFT~\cite{openvlaoft} & 56.4 & 31.9 & 79.5 & 88.7 & 93.3 & 75.8 & \textbf{74.2} & \textbf{69.6} \\
OpenVLA-OFT+$\dagger$ (plus mix-sft) & 92.8 & 30.3 & 85.8 & 94.9 & 93.9 & 89.3 & 77.6 & 79.6 \\
\bottomrule
\end{tabular}
\end{table*}

\subsection{Official-Protocol Perturbation Evaluation}
\label{sec:pro}

\begin{figure*}[t]
\centering
\includegraphics[width=0.8\textwidth]{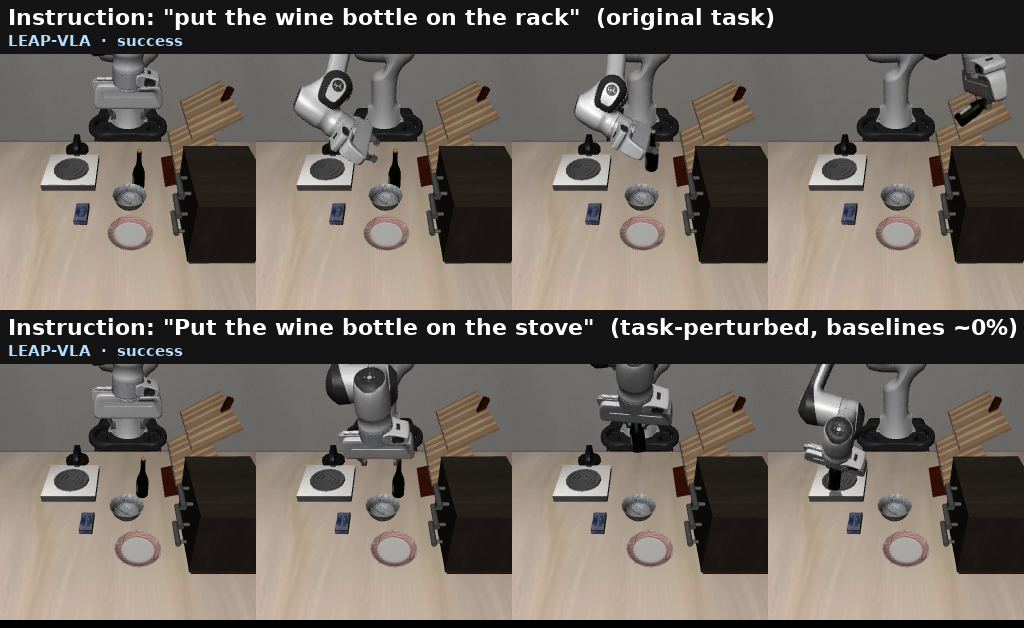}
\caption{Task-perturbation generalization on LIBERO-PRO (libero\_goal).
Top: the original task ``put the wine bottle on the rack'', which appears
in the training data---our model completes it. Bottom: only the goal is
changed (``Put the wine bottle on the stove''); the scene and objects are
identical. Our model still executes the new instruction, whereas VLA
baselines score 0.0--6.8\% on this perturbation dimension (ours: 20.4\%
averaged over the four suites, 8.4--49\% per suite). Both rows are
successful episodes re-recorded with the evaluation protocol of
Table~\ref{tab:pro} (asynchronous deployment: semantic anchor refreshed
at each 0.5\,s re-plan, fast stream and measured cache age updated every
control step); frames are keyframes sampled from the 100\,ms-interval
agentview replay.}
\label{fig:pro-task}
\end{figure*}

We additionally evaluate on LIBERO-PRO~\cite{liberopro}, the
official-protocol perturbation evaluation of the LIBERO suites: each
suite in its original form plus five perturbation dimensions (object swap,
position swap, language, task, environment), evaluated at the official
protocol's sample size of 50 initial states per perturbed task, i.e.,
500 episodes per suite--dimension. The checkpoint is the same four-suite LIBERO model at
600k steps, again trained only on the original LIBERO suites and served
by the asynchronous deployment server under the deployed-age protocol of
Sec.~\ref{sec:setup} (fixed 0.5\,s re-plan clock, the semantic condition
refreshed at every re-plan). Table~\ref{tab:pro} reports the
final cross-dimension summary against the official baselines. Two readings
stand out: the task dimension (20.4\%) is far above every baseline
(0.0--6.8\%), while position swap collapses to
9.5\%, within the 0.0--20.8\% baseline range on this dimension.
Figure~\ref{fig:pro-task} visualizes the task-dimension gap with strictly
paired episodes (same scene, same initial layout, only the goal changed);
Appendix~\ref{app:qual} complements it with environment- and
language-dimension pairs and a failure-mode analysis.

\begin{table*}[t]
\centering
\caption{Official-protocol perturbation summary: success (\%) averaged
over the four LIBERO suites per dimension. Baseline rows are quoted from
the official evaluation; ``---'' marks dimensions not reported for that
model (excluded from its average). \method{} is trained only on the
original LIBERO suites (Qwen2.5-VL-2B backbone) and evaluated zero-shot on every
perturbation; perturbation initial states are regenerated by our harness
(same distribution).}
\label{tab:pro}
\begin{tabular}{lcccccc}
\toprule
Model & Object & Position & Language & Task & Environment & Avg \\
\midrule
$\pi_{0.5}$~\cite{pi05} & 96.0 & 20.8 & 95.8 & 0.8 & 52.8 & \textbf{53.2} \\
OpenVLA~\cite{openvla} & 93.0 & 0.0 & 97.3 & 0.0 & 68.0 & 51.7 \\
\method{} (ours)  & 90.0 & 9.5 & 85.7 & 20.4 & 45.1 & 50.1 \\
X-VLA~\cite{xvla} & 79.0 & 0.8 & 96.8 & 6.8 & --- & 45.8 \\
$\pi_0$~\cite{pi0} & 90.5 & 0.0 & 90.5 & 0.0 & 38.8 & 44.0 \\
Molmoact~\cite{molmoact} & 76.0 & 1.5 & 85.8 & 1.5 & --- & 41.2 \\
\bottomrule
\end{tabular}
\end{table*}

\subsection{Real-Robot Results}
\label{sec:real}

\begin{figure*}[t]
\centering
\includegraphics[width=0.9\textwidth]{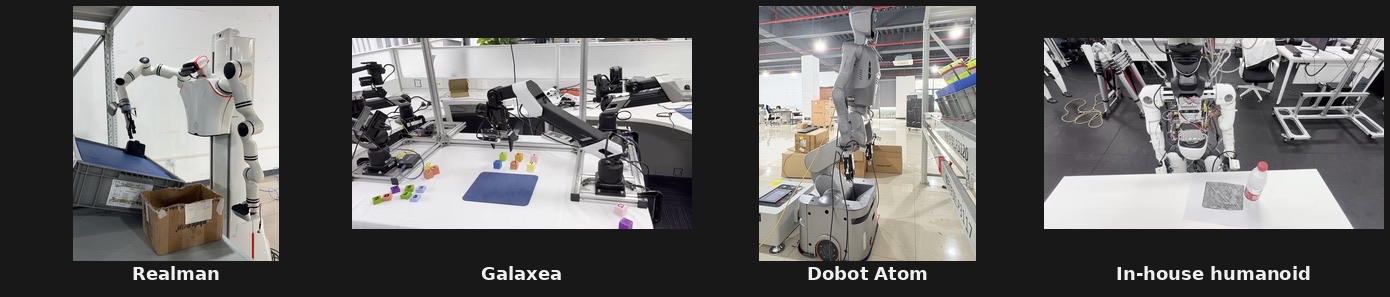}
\caption{Real-robot deployment platforms. From left to right: Realman
dual-arm workstation, Galaxea dual-arm workstation, Dobot Atom mobile
dual-arm humanoid, and our in-house humanoid upper body. All policies run
on-board on the same \method{} checkpoint layout; only the embodiment
config differs per platform.}
\label{fig:platforms}
\end{figure*}

We deploy \method{} on four platforms (Fig.~\ref{fig:platforms}): two
dual-arm workstations (Realman, Galaxea), the Dobot Atom mobile dual-arm
humanoid, and our in-house humanoid upper body. All tasks run from the same
onboard camera layout as training, with no per-task tuning;
Table~\ref{tab:real} reports per-task success.

\begin{table}[t]
\centering
\caption{Real-robot success rate, 50 trials per task. Tasks correspond to
the seven rows of Fig.~\ref{fig:real-rollouts}. Both columns use the
\method{} architecture (Sec.~\ref{sec:method}): Same-freq.\ is trained
without staleness awareness and deployed with the VLM at every control
step (the compute-unconstrained anchor); \method{} is the full
staleness-aware recipe deployed under slowed semantic refresh. The naive-reuse collapse that motivates the
recipe is quantified in simulation (Sec.~\ref{sec:deploy}).}
\label{tab:real}
\setlength{\tabcolsep}{4pt}
\begin{tabular}{lcc}
\toprule
Task & our same-freq. & \method{} \\
\midrule
\multicolumn{3}{l}{\emph{Realman}} \\
Sort blocks by color          & 0.94 & 0.94 \\
Pack workpieces into box      & 0.74 & 0.70 \\
4-piece pick-and-place        & 1.00 & 1.00 \\
\addlinespace
\multicolumn{3}{l}{\emph{Dobot Atom}} \\
Pick-and-place workpiece      & 0.86 & 0.80 \\
\addlinespace
\multicolumn{3}{l}{\emph{Galaxea}} \\
Sort by color                 & 1.00 & 1.00 \\
Word spelling                 & 0.94 & 0.90 \\
\addlinespace
\multicolumn{3}{l}{\emph{In-house humanoid}} \\
Bottle placement              & 0.96 & 0.96 \\
\bottomrule
\end{tabular}
\end{table}

\textbf{Protocol.} Each task is evaluated for 50 trials on its platform.
A trial succeeds when the task's end condition holds: blocks grouped into
two color-pure piles; the carton packed and closed; all four workpieces in
the side cartons; the workpiece resting in the target bin; colored
workpieces sorted with the distractor left out; the two letter rows
completed; and the bottle standing upright inside the marked target zone.
Same-freq.\ runs the full VLM at every control step; \method{} runs the
asynchronous stack of Sec.~\ref{sec:deploy}---semantic refresh every
four control steps ($1{:}4$), with the fresh fast stream and the measured
cache age updated at every control step.
In the interactive task a human only hands over and re-stands the bottle
between autonomous cycles; trials are scored on the autonomous segments.

\begin{figure*}[t]
\centering
\includegraphics[width=\textwidth]{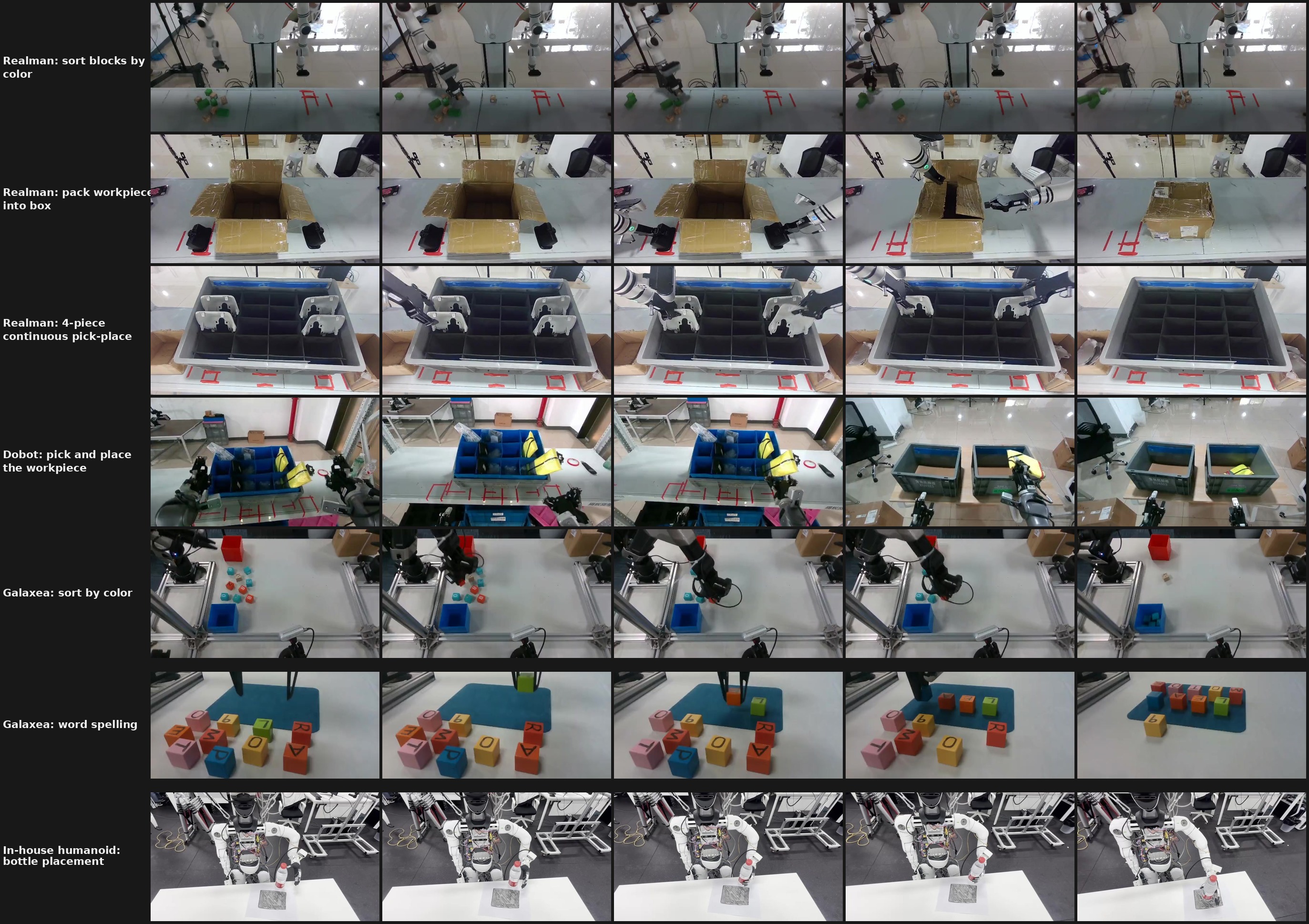}
\caption{Qualitative real-robot rollouts on the four platforms of
Fig.~\ref{fig:platforms}. Each row shows five keyframes of one episode
(start, approach, grasp, transport, final state) sampled from the onboard
camera stream, one task per row: sorting blocks by color and packing
workpieces into a box (Realman, rows 1--2); four-piece continuous dual-arm
pick-and-place (Realman, row 3); pick-and-place of a workpiece (Dobot Atom,
row 4); color-based sorting and word spelling with letter cubes (Galaxea,
rows 5--6); placing a bottle into a marked target zone (in-house humanoid,
row 7). All keyframes are sampled from successful autonomous segments of the
policy rollout; in the interactive task (row 7) a human only hands over and
re-stands the bottle between cycles, which is not shown.}
\label{fig:real-rollouts}
\end{figure*}

% ============================================================================
\section{Do VLAs Know When the Scene Doesn't Match the Instruction?}
\label{sec:mismatch}

\begin{figure*}[t]
\centering
\begin{tabular}{@{}c@{\hspace{6pt}}c@{}}
\includegraphics[width=0.49\textwidth]{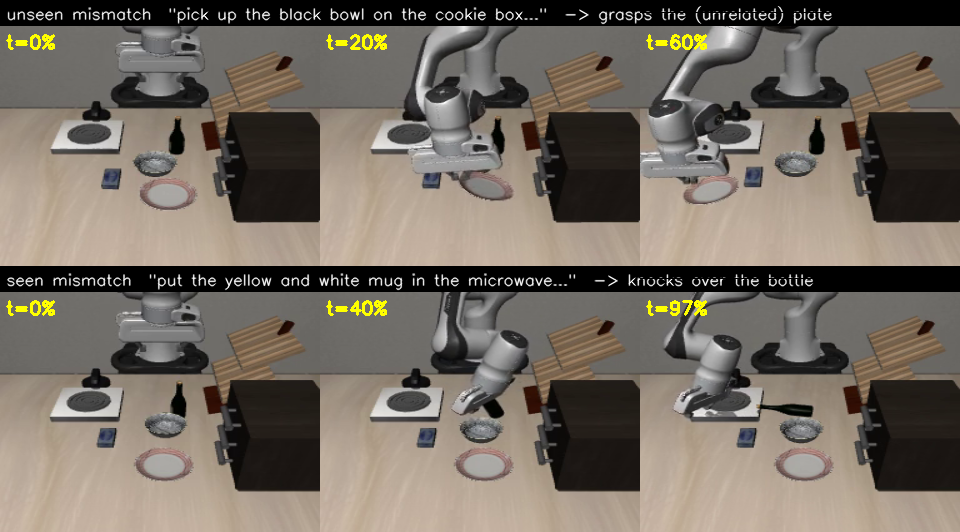} &
\includegraphics[width=0.49\textwidth]{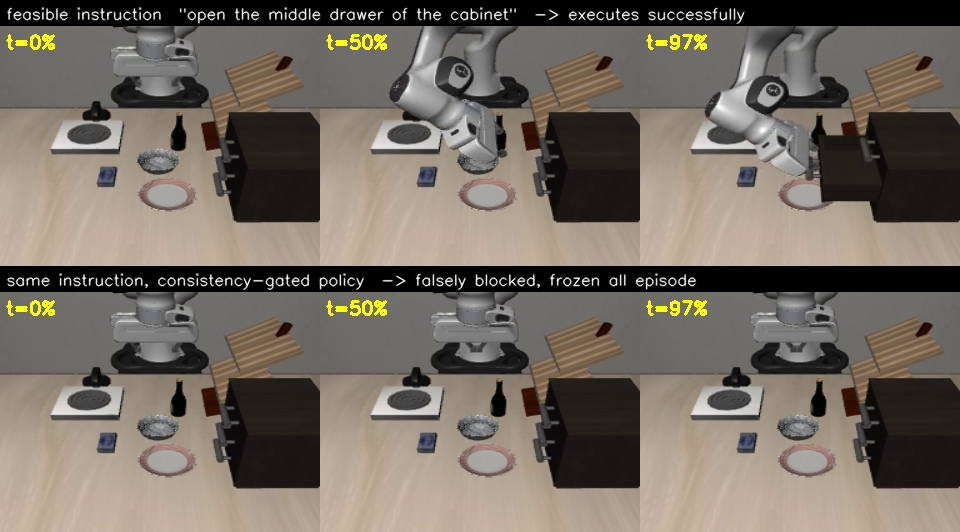}
\end{tabular}
\caption{Left: false execution under instruction--scene mismatch---each
row is one rollout at increasing time fractions, instruction quoted
above; the instruction refers to absent objects, yet the policy grasps
an unrelated plate (top) or knocks over a bottle (bottom). Right: a
consistency-gated policy freezes on a feasible instruction (``open the
middle drawer of the cabinet'', t=0/50/97\%): the negatives-only policy
executes correctly, while behind a CoT-verdict gate misjudged as
inconsistent the policy freezes for the entire episode (0/20 feasible
tasks under gating vs 6/20 without)---a property of scene-imbalanced
gate supervision, not of gating itself: after scene-balanced retraining
the gate passes 20/20 feasible tasks and matches the ungated policy.}
\label{fig:mismatch}
\end{figure*}

Every VLA in this paper's lineage is trained the same way, and the way is
itself a defect: the training distribution consists entirely of successful
demonstrations, so the policy is never asked whether the instruction is
\emph{satisfiable} --- $P(\text{action}\mid \text{obs}, \text{instr})$ is
sampled only where it is --- and the action head has no idle or abstain
mode. The result is an \emph{achievability prior}: under an
instruction--scene mismatch the policy does not idle, it keeps manipulating
the scene --- grasping unrelated objects that happen to be present,
knocking over obstacles (Fig.~\ref{fig:mismatch}) --- because the
language-to-action shortcut bypasses visual verification altogether. This
is the same training-side gap that leaves the leftover conflict of
Appendix~\ref{app:conflict} unresolved: in both cases the supervision
never contains a single sample whose correct response to ``the observation
disagrees with what the instruction (or prefix) says'' is to stop.

\textbf{Zero-cost negatives.} We close part of the gap with a data
intervention that costs no extra collection: re-assign instructions from
the other LIBERO suites to goal-scene episodes and keep a pair only when
at least one referenced object is absent (the simulator's bddl object
list acts as the geometric ground truth), labeled with the zero action
(idle). This yields 240 episodes over 27 mismatched prompts (120 pairs),
23\% of fine-tuning exposure, trained for 15k steps from the same base
checkpoint --- no teleoperated failures, no environment modification, no
new loss terms.

\textbf{CoT annotations.} Keyframes sampled from these episodes are
annotated; each carries a short structured rationale: the task, the
current subtask, the next step, a scene description, and the grounded
object boxes, generated offline by a VLM annotation pipeline. The
scene description always reflects the observed frame, so under an
instruction--scene mismatch it simply does not contain the referenced
object---this is where the mismatch signal enters supervision.

\textbf{Protocol and metric.} Mismatch modes: 30 pairs $\times$ 3
trials (90 episodes each); feasible: 10 tasks $\times$ 2 trials; 300
steps, EMA weights on the f32 deployment stack. Mismatch episodes that
stay still for 100 consecutive steps end early (frozen rollouts only;
per-frame motion statistics are unaffected). Environment success is
undefined on mismatched tasks (the referenced objects are not present),
so behavior is the metric: rollout motion (mean absolute grayscale
frame difference, 0--255 scale) and the fraction of active frames
(difference $>$1.0); for calibration, a fully frozen rollout reads
0.02/0\%.

\begin{table*}[t]
\centering
\caption{Behavior under instruction--scene mismatch (mismatch modes: 30
pairs $\times$ 3 trials; feasible: 10 tasks $\times$ 2 trials; 300
steps). Mismatch cells report rollout motion
(mean absolute grayscale frame difference) / fraction of active frames;
a fully frozen rollout reads 0.02/0\%. $^{*}$Qualitative probe: the
success-only policy keeps manipulating the scene on essentially every
mismatched rollout (Fig.~\ref{fig:mismatch}); motion metrics were not
run for it.}
\label{tab:mismatch}
\begin{tabular}{lccc}
\toprule
Config & Seen mismatch & Unseen mismatch & Feasible success \\
\midrule
Success-only baseline & $\approx$100\% false execution$^{*}$ & $\approx$100\% false execution$^{*}$ & -- \\
$+$ zero-cost negatives & 1.29\,/\,41\% & 1.31\,/\,41\% & \textbf{20/20} \\
$+$ goal-scene positives (no CoT) & 2.24\,/\,66\% & 2.28\,/\,70\% & \textbf{20/20} \\
$+$ positives $+$ CoT verdict (policy only) & 0.05\,/\,0\% & 0.05\,/\,0\% & \textbf{20/20} \\
$+$ positives $+$ CoT verdict (gated) & \textbf{0.02\,/\,0\%} & \textbf{0.02\,/\,0\%} & \textbf{20/20} \\
\bottomrule
\end{tabular}
\end{table*}

Table~\ref{tab:mismatch}: negative supervision cuts the fraction of
active frames on mismatched rollouts to ${\sim}40\%$ at zero regression
on feasible tasks (20/20). The honest reading is ``reduces, does not
eliminate'': motion is far from the frozen floor, and roughly four in
ten mismatch episodes still intervene with the scene --- similarly on
held-out pairs, so the deficit is uniform rather than a
generalization gap. The asymmetry is
structural: in flow matching the ``execute the matching instruction''
behavior is a dense target manifold while ``exactly zero action'' is a
thin one, so leaving the manifold is easy but hitting zero everywhere is
not (Fig.~\ref{fig:mismatch} documents the residual: unrelated-object
grasps and obstacle knocks on mismatched instructions). Adding
goal-scene positives without CoT supervision does not close the gap ---
it dilutes the negative exposure and false execution worsens further
(rows~2--3). What closes it is the chain-of-thought verdict: with CoT
added, the policy itself reaches the frozen floor on the active-frames
metric for every mismatched rollout (0\%, all videos static) while still
executing feasible tasks (20/20, matching the CoT-free rows at the final
auxiliary-loss weight).

\textbf{Why the verification head needs balanced supervision.} We also
trained a parallel chain-of-thought verdict head on the same data
(deployment-side gating on its verdict). Its first version fails
informatively. (1)~\emph{Scene shortcut}: trained on scene-imbalanced
supervision --- all negative verdicts on goal scenes, all positives
elsewhere --- the binary head learned scene identity, not
instruction--scene consistency: it blocked every rollout, feasible
tasks included, with hallucinated rationales citing objects absent from
the instruction, while the flow-matching action head trained on the same
data learned no such shortcut (20/20 feasible).
(2)~\emph{Interference}: the auxiliary LM loss at weight 1.0 degraded the
action head through their shared trainable parameters (feasible 20/20
$\to$ 6/20 without the gate), so head isolation or a much smaller weight
is required (Fig.~\ref{fig:mismatch}, right).

Both failure modes are addressed by construction, not by architecture:
mixing the goal suite's own (feasible) instructions into the CoT
supervision on the \emph{same} goal scenes makes scene identity
uninformative for the verdict, and a loss-weight sweep (1.0, 0.2, 0.1,
0.05) localizes the interference: mismatch suppression is identical at
every weight (frozen floor, 0\% active frames), while feasible success
rises monotonically as the weight decreases (6/20 $\to$ 16/20 $\to$
19/20 $\to$ 20/20) --- the cost is a pure loss-weight artifact, not an
inherent price of CoT supervision. At the final weight (0.05) the
verdict head blocks 166/180 mismatched episodes at the 30-pair scale
(85/90 on seen pairs, 81/90 on unseen) and passes 20/20 feasible tasks;
across the 180 generated rationales, every blocked verdict names the
actual absent objects and every passed verdict affirms presence ---
no hallucinations. All misses fall in one borderline family (a
referenced object partially present) whose verdict flips across trials
--- and whose rollouts are nonetheless frozen by the policy itself, so
the two layers are redundant rather than fragile. Under the gate,
feasible-task behavior is identical to the ungated policy (same 20/20,
same successes), i.e.\ pass-through costs nothing, and at the final
weight the auxiliary loss itself costs nothing either.

Scope: one model family, one benchmark, and a frame-difference proxy for
manipulation (mismatch modes: 30 pairs $\times$ 3 trials; feasible: 10
tasks $\times$ 2 trials). The numbers justify the negative-data
construction and the scene-balance + weight recipe for the verdict head,
which at the final weight leaves feasible success untouched; the open
item is a discrete behavioral metric to replace the frame-difference
proxy.

% ============================================================================
\section{Limitations}
\method{} trains and evaluates staleness up to $\Delta_{\max}$; ages beyond
the trained range are not covered, and triggering-based refreshes trade
compute for smaller effective ages rather than extending the range. Even
within the trained range, measured-age conditioning and the corruption
curriculum recover deployed success but do not create information:
between refreshes the system holds no representation of what has happened
since. Inference-time boundary corrections
(Sec.~\ref{sec:smooth}) smooth the deployment-typical self-leftover regime
but cannot force execution to follow a leftover that genuinely conflicts
with the fresh observation (Appendix~\ref{app:conflict}); closing that gap
requires training-side prefix-conflict adaptation, and our initial attempts
(Appendix~\ref{app:crossckpt}) did not close it. Finally, our evaluation is
confined to quasi-static simulation suites and a small real-robot set;
broader dynamic benchmarks
are needed to substantiate the reactivity claims at scale. The fast-stream
encoder is frozen; end-to-end finetuning of it---and the drift this would
cause in the learned stale-versus-fresh arbitration---is unexplored.

% ============================================================================
\section{Conclusion}
\method{} turns a VLA's semantic backbone from a per-step bottleneck into a
low-frequency anchor, and pays for the resulting staleness explicitly: the
expert is conditioned on the measured age of its semantic cache and trained
over the full staleness distribution, while the fresh visual stream---motion-aware
image tokens---enters the expert through one
pluggable token-block interface and carries the reactive burden at control
rate. The result is a policy that
preserves its fresh anchor, matches same-frequency inference, and scales to
backbones and hardware where joint-frequency inference is infeasible. The
deployed form of the stack is fully forward-only: staleness-aware training
carries the distribution, a gradient-free boundary correction carries the
chunk seams, and neither pays an autograd or latency tax, so the control
path exports to compiled runtimes unchanged.

\textbf{Scope and transfer.} This is an architecture-level study: it asks
what a shared-attention VLA loses when its semantic loop slows, and answers
with single-variable training measurements rather than leaderboard entries,
so comparisons are restricted to the VLA class under one deployment
protocol. World-action models, which couple action generation to predicted
future world states, optimize a different axis and are not compared. The
dual-rate systems of Sec.~\ref{sec:related} compose separately trained
models whose fast side consumes the slow one only through language or a
fixed latent, and none characterizes the
staleness distribution its fast loop faces---the gap the ladder of
Sec.~\ref{sec:deploy} measures directly. The ingredients are portable by
construction: asynchronous-$\Delta t$ exposure, measured-age conditioning,
the corruption curriculum, and soft-nudge modify only the training
distribution and the inference-time action update, so they apply to any VLA
that re-runs its backbone at every re-planning step---demonstrated here
at a 2B semantic backbone (the compared rows span $0.9$--$7$B). We believe
measured-age conditioning is a general tool for any architecture that reuses
context across control steps, and we release our asynchronous evaluation
protocol to make such evaluations standard.

% ============================================================================

\clearpage

% ============================================================================
\appendices
\section{Asynchronous Evaluation Protocol}
\label{app:protocol}

All deployment-line measurements
(Tables~\ref{tab:ladder}--\ref{tab:smooth}) run the same physical-clock
loop, which reproduces the deployed queue semantics in simulation. The
client executes one action per 50\,ms. Control steps are grouped into re-planning
windows of $R{=}10$ (0.5\,s; $R{=}1$ in the $0.05$\,s columns of
Table~\ref{tab:deploy}); within an $R{=}10$ window, the observation for the next request
is captured at step $K{=}6$ and the request is issued, carrying the
\emph{leftover}: the not-yet-executed tail of the currently executing
chunk, in raw action space, together with the number of already-executed
steps. When the window ends, the returned chunk is adopted with its first
$R{-}K{=}4$ steps discarded---exactly the control steps that elapse while
the chunk is computed---and the remaining steps execute over the next
window. The first window has no leftover. The consumption schedule is
bit-identical across all compared corrections; the only difference between
systems is what the server does with the leftover.

\textbf{Staleness accounting.} Every cell records, per request, whether it
took a fresh context or reused the cache, and the age of the context it
used. These sentries must match the scheduled regime exactly---refresh
fraction $1.00$ and cache age ${\equiv}\,0$ at the matched anchors (the
DiT:VLM $1{:}1$ operating point of Sec.~\ref{sec:setup}); the alternating/cycling
age patterns of Table~\ref{tab:ratio} at the $2{:}1$/$4{:}1$ cells---and every
reported cell passes this check.

\begin{figure}[t]
\centering
\includegraphics[width=\linewidth]{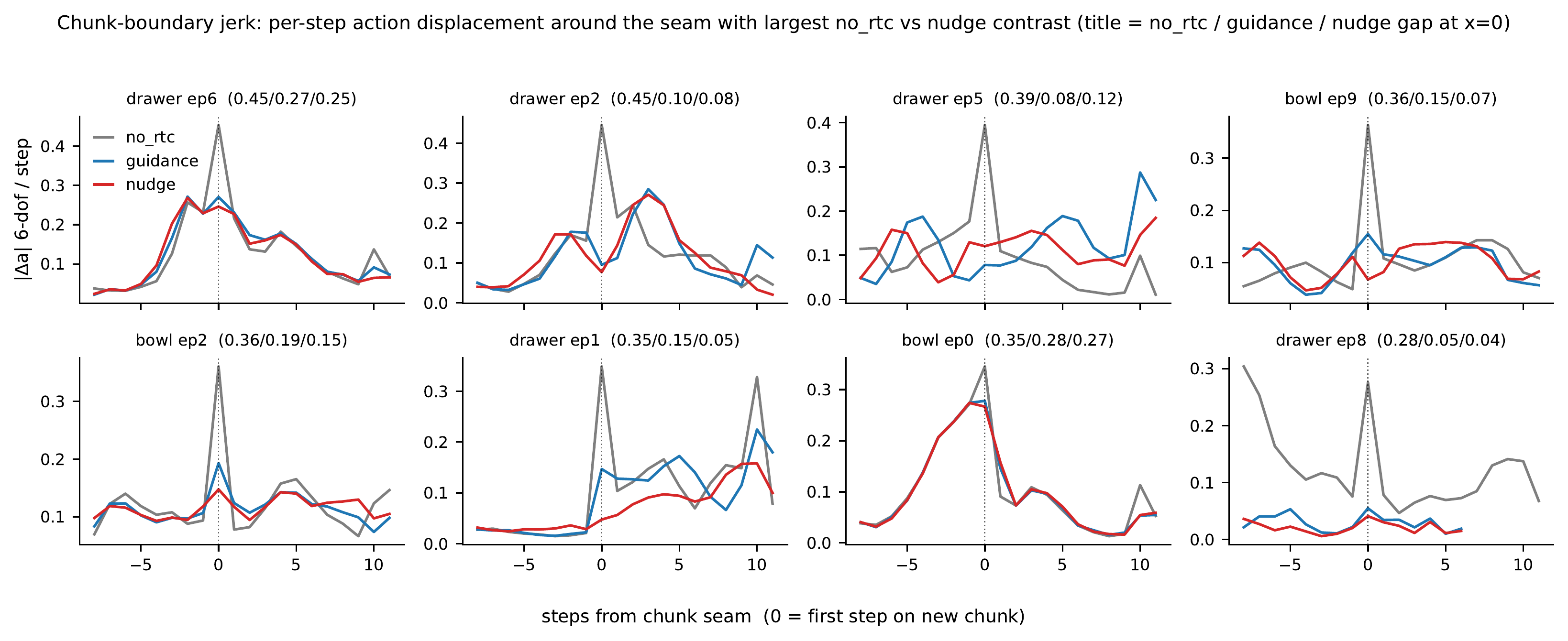}
\caption{The eight boundary windows with the largest
uncorrected-vs-corrected contrast, one per episode (two LIBERO-Goal tasks,
200-trajectory evaluation; Table~\ref{tab:smooth}). Each panel shows the
per-step 6-DoF displacement around one boundary (step 0 = first executed
step on the new chunk; each method aligned to its own nearest boundary);
the title gives the boundary displacement of all three executions
(no-correction/guidance/nudge). Both corrections remove
most of the boundary spike; soft-nudge consistently more.}
\label{fig:seam-windows}
\end{figure}

\begin{figure}[t]
\centering
\includegraphics[width=\linewidth]{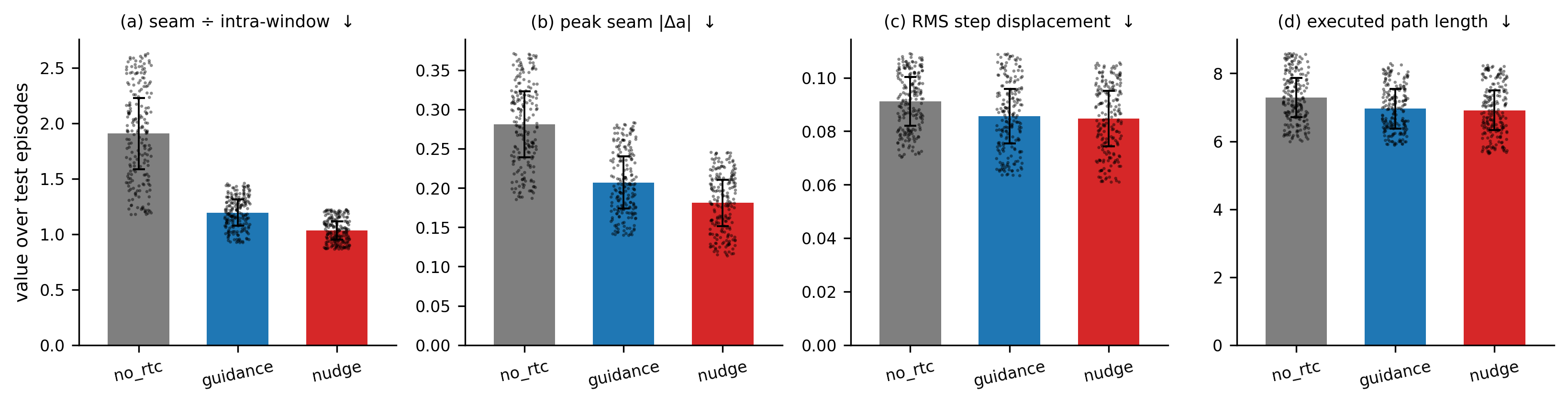}
\caption{Per-episode distributions of the four smoothness metrics of
Table~\ref{tab:smooth} (bars: mean $\pm$ 95\% CI; dots: individual
episodes).}
\label{fig:metric-bars}
\end{figure}

\textbf{Boundary metric.} For each adopted chunk we record the 6-DoF
displacement norm of its first executed step (the boundary step) and the
intra-window mean displacement; their ratio (boundary/intra) aggregates per
episode. RTC
guidance uses the exponential prefix schedule with execution horizon
$H{=}10$ and maximum guidance weight $\gamma{=}10$; soft-nudge uses the
same schedule with $\eta{=}1$ (Eq.~\ref{eq:nudge}); Figs.~\ref{fig:seam-windows}
and~\ref{fig:metric-bars} show the highest-contrast windows and the
per-episode distributions.

\section{Leftover-Conflict Geometry}
\label{app:conflict}

Soft-nudge and RTC guidance both pull execution toward the leftover prefix.
This appendix measures how much pulling is possible as a function of what
the leftover \emph{is}. We inject leftovers of two provenances into the
base checkpoint's denoising loop and compare hard inpainting (leftover rows
clamped, as in RTC's inpainting variant), autograd guidance, and soft-nudge
at $\eta \in \{0.1, 0.3, 1.0\}$, all with the same prefix schedule. The
\emph{stale} leftover is the policy's own just-computed plan, offset in
time so that it conflicts with the fresh observation (conflict
$\lVert c \rVert {=} 0.166$); the \emph{foreign} leftover comes from a
different episode ($\lVert c \rVert {=} 1.237$), off the manifold of the
current context. Table~\ref{tab:sensitivity} reports the resulting
distances. At the first row executed after the frozen prefix (the
conflict row) we measure $d_{\mathrm{tail}}$, the residual distance between
the executed action and the leftover row aligned to the same control step;
the bridging fraction $1 - d_{\mathrm{tail}} / \lVert c \rVert$; and
$d_{\mathrm{gt}}$, the tracking error of the executed suffix against the
ground-truth trajectory (the no-leftover reference attains 0.080).

\begin{table*}[t]
\centering
\caption{Leftover-conflict probe (base checkpoint, 8 paired samples).
Left: stale self-leftover; right: foreign (off-manifold) leftover.
Bridging $= 1 - d_{\mathrm{tail}}/\lVert c\rVert$. No correction reaches
the leftover at the conflict row; suffix tracking is preserved throughout.}
\label{tab:sensitivity}
\begin{tabular}{lccccc}
\toprule
& \multicolumn{2}{c}{Stale self-leftover ($\lVert c\rVert = 0.166$)} &
  \multicolumn{3}{c}{Foreign leftover ($\lVert c\rVert = 1.237$)} \\
\cmidrule(lr){2-3}\cmidrule(lr){4-6}
Correction & $d_{\mathrm{tail}}$ & $d_{\mathrm{gt}}$
           & $d_{\mathrm{tail}}$ & bridging & $d_{\mathrm{gt}}$ \\
\midrule
None (reference)        & --    & 0.080 & --    & --            & 0.080 \\
RTC inpainting~\cite{rtc} & 0.170 & 0.086 & 1.204 & 2.7\%         & 0.085 \\
RTC guidance~\cite{rtc}   & 0.164 & 0.082 & 1.208 & 2.4\%         & 0.092 \\
Soft-nudge $\eta = 0.1$   & 0.165 & 0.081 & 1.235 & 0.2\%         & 0.080 \\
Soft-nudge $\eta = 0.3$   & 0.165 & 0.081 & 1.229 & 0.7\%         & 0.079 \\
Soft-nudge $\eta = 1.0$ (used) & 0.164 & 0.082 & 1.205 & 2.6\%   & 0.093 \\
\bottomrule
\end{tabular}
\end{table*}

Three readings. \textbf{(1) Conflicts are not forceable.} On stale
self-leftovers, every mechanism---including clamping the prefix rows
outright---leaves the conflict row at essentially the conflict norm
($d_{\mathrm{tail}} \approx \lVert c \rVert$) while suffix tracking stays
at the no-leftover level: the policy departs smoothly from a stale plan
rather than following it, and no inference-side force changes that. The
boundary benefit of Sec.~\ref{sec:smooth} therefore lives in the
deployment-typical regime---a leftover the policy itself just produced,
roughly consistent with the current observation---not in forcing execution
onto plans the evidence has outgrown. \textbf{(2)~Forward-only
equivalence.} On adversarial (foreign) leftovers, where corrections have
their maximal effect, soft-nudge at $\eta{=}1$ bridges $2.6\%$ of the
conflict against $2.4\%$ for autograd guidance at its deployed strength,
with equal suffix error: the identity-Jacobian substitution of
Eq.~\ref{eq:nudge} is behaviorally a stand-in for guidance.
\textbf{(3)~Strength saturates, then taxes.} Even at maximum strength,
bridging never exceeds $2$--$3\%$ for any mechanism, and at full strength
both guidance and
soft-nudge begin to tax suffix tracking on adversarial leftovers
($d_{\mathrm{gt}}$ $0.080 \to 0.092$/$0.093$). The residual conflict is a
training-side gap---the base model has never been asked to continue a
prefix that disagrees with its observations---and must be closed in
training, not by pushing inference harder.

\section{Consistency Across Checkpoints and Negative Attempts}
\label{app:crossckpt}

\begin{table*}[t]
\centering
\caption{Boundary/intra ratio without and with inference-time correction,
across checkpoint families. Rows use different suites, episode sets, and
window sizes; compare within a row only.}
\label{tab:crossckpt}
\begin{tabular}{lccc}
\toprule
Checkpoint family & No corr. & RTC guidance & Soft-nudge \\
\midrule
Frozen-backbone expert (Goal) & 2.87 & 1.04 & -- \\
Deployed expert (Goal) & 1.64 & 1.11 & -- \\
Deployed expert $+$ prefix fine-tuning (15k) & 2.05 & 1.04 & -- \\
Deployed expert (Table~\ref{tab:smooth} protocol) & 1.91 & 1.19 & \textbf{1.04} \\
\bottomrule
\end{tabular}
\end{table*}

The boundary reduction of Sec.~\ref{sec:smooth} is not an artifact of one
checkpoint: Table~\ref{tab:crossckpt} reproduces it on four checkpoint
families spanning two architecture generations and two training recipes,
with guidance reducing the boundary ratio on every family and soft-nudge
matching or exceeding it where all three were run.

\textbf{Negative attempts at training-side prefix-following.} We also tried
to move prefix-following into training, and record both failures. First,
fine-tuning the deployed expert for 15k steps on prefix-continuation
targets (the first $d$ rows of a ground-truth trajectory as prefix, the
rest as prediction target; row 3 of Table~\ref{tab:crossckpt}) does not
teach it: evaluated without correction the boundary ratio is 2.05, and
executing its chunks by inpainting reduces boundaries no better than
discarding the leftover (1.94 vs.\ 2.05), while guidance-style correction
on the same checkpoint still reaches 1.04. A prefix-ablation probe shows
the gap is not \emph{reading}: the base model already conditions on clean
prefixes (replacing them with noise or foreign rows raises the suffix
loss), and fine-tuning strengthens this by only ${\sim}30\%$; what training
never supplies is a signal for which side to take when the prefix and the
fresh observation disagree, because training prefixes are always consistent
with their own trajectory. Second, we added exactly that signal as a
conflict augmentation---prefixes replaced with smooth staleness-drift
curves at 35\% of prefix steps, plus a boundary-smoothing auxiliary
loss---and trained for 6k steps: bridging of drifted leftovers, read out by
the probe of Appendix~\ref{app:conflict}, did not move at any checkpoint
snapshot. Lightweight fine-tuning does not overturn an
observation-dominated prior; closing the residual conflict of
Appendix~\ref{app:conflict} remains an open training-side problem
(leftover-gated adaptation).

\section{Qualitative Results on Perturbation Benchmarks}
\label{app:qual}

\begin{figure*}[t]
\centering
\includegraphics[width=\textwidth]{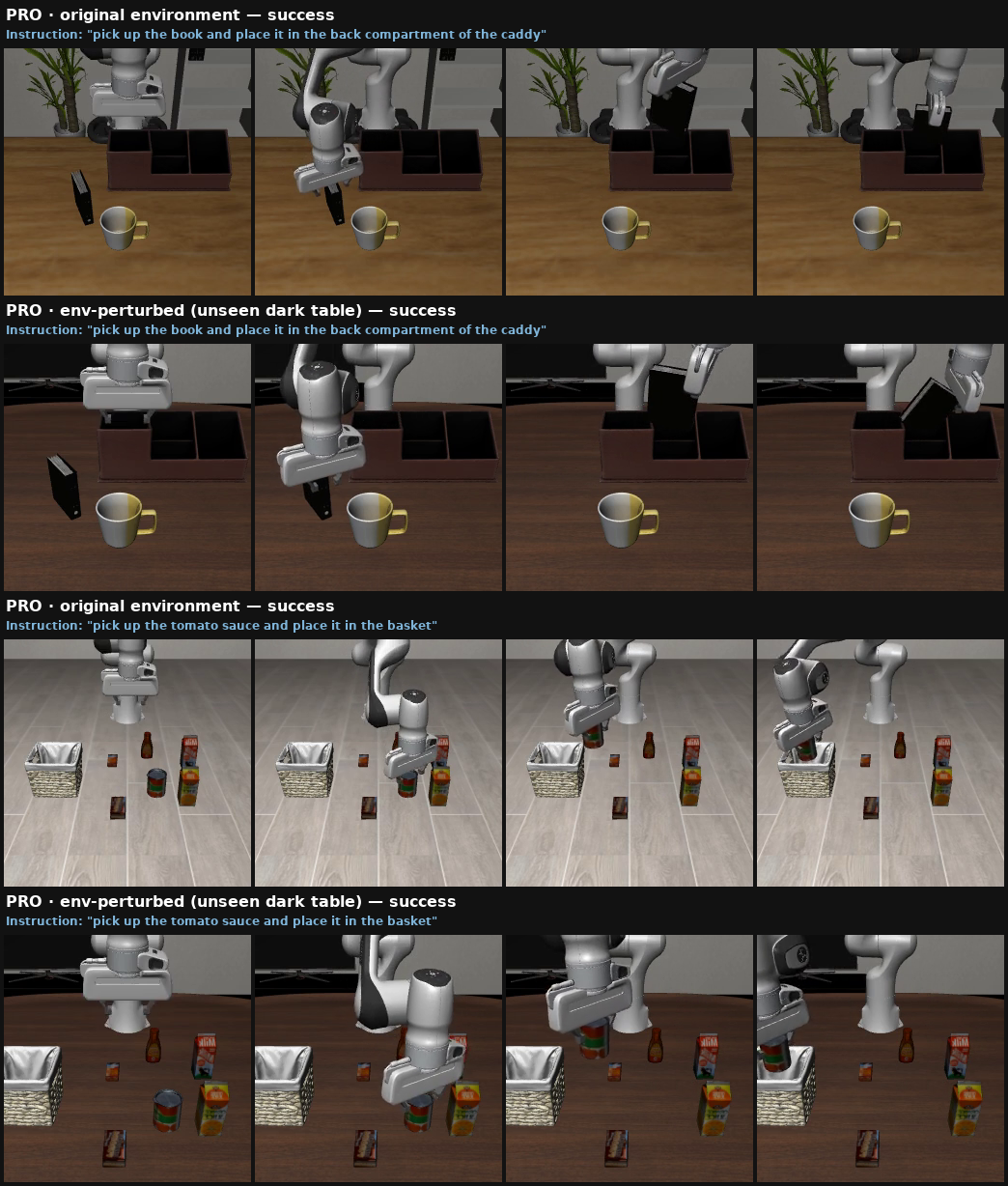}
\caption{Environment-perturbation generalization on LIBERO-PRO. Each pair
shows the same task in the original environment (top) and in an unseen
perturbed environment (bottom): the light table is replaced by a dark
wooden one, with background and lighting changed as well. Our model
completes the task in both settings---placing the book into the rear
compartment of the caddy (rows 1--2) and the tomato sauce into the basket
(rows 3--4). Episodes are successful takes re-recorded with the
evaluation protocol of Table~\ref{tab:pro}; frames are keyframes of the
100\,ms-interval agentview replay.}
\label{fig:pro-env}
\end{figure*}

\begin{figure*}[t]
\centering
\includegraphics[width=\textwidth]{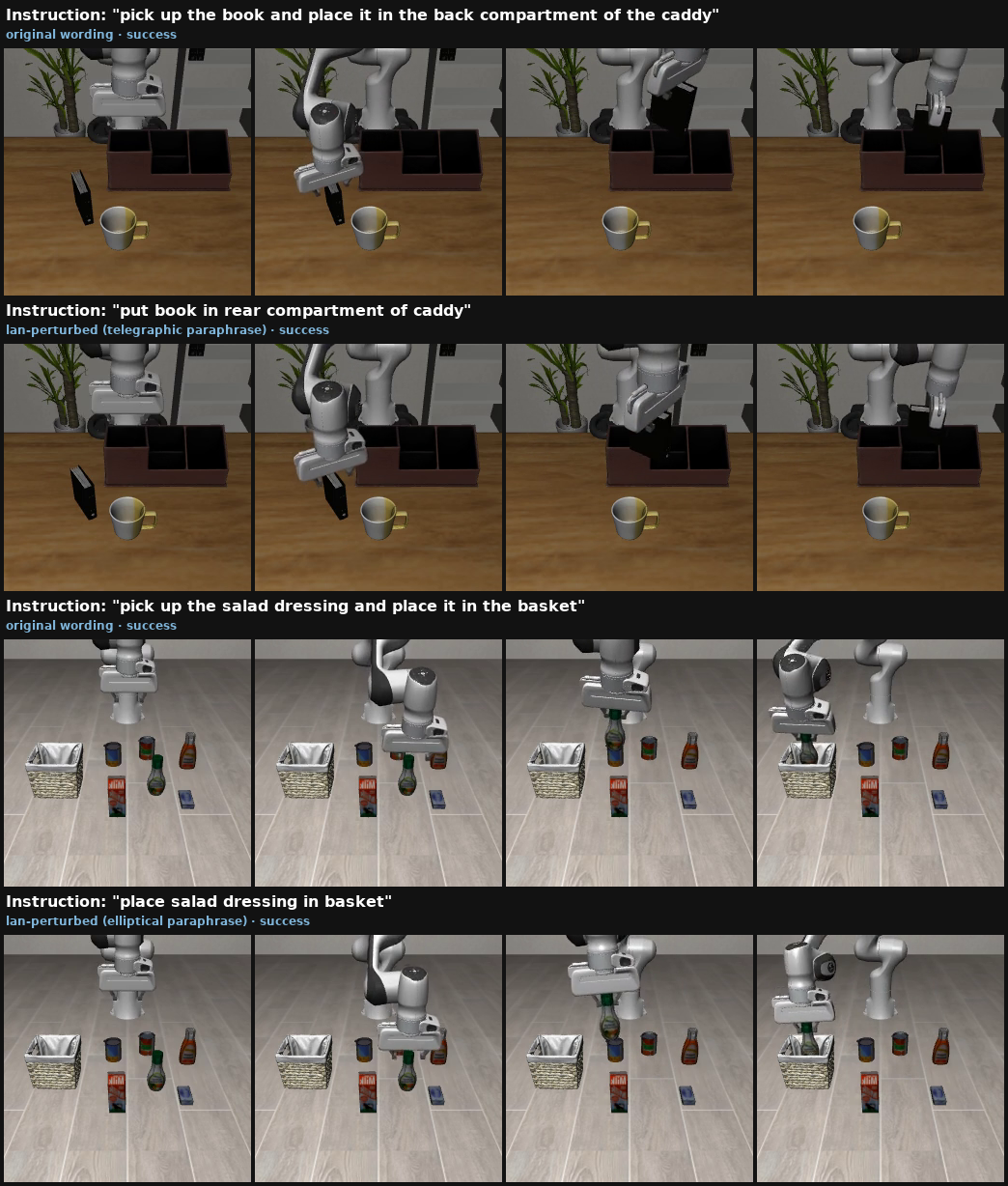}
\caption{Language-perturbation robustness on LIBERO-PRO. Each pair shows
the same task from the same initial state under the original instruction
(top) and a paraphrase (bottom): a telegraphic rewording (``put book in
rear compartment of caddy'', rows 1--2) and an elliptical one (``place
salad dressing in basket'', rows 3--4). The executed behavior is
essentially unchanged and both episodes succeed, indicating the policy
follows the task semantics rather than surface word patterns.}
\label{fig:pro-lan}
\end{figure*}

\begin{figure*}[t]
\centering
\includegraphics[width=0.97\textwidth]{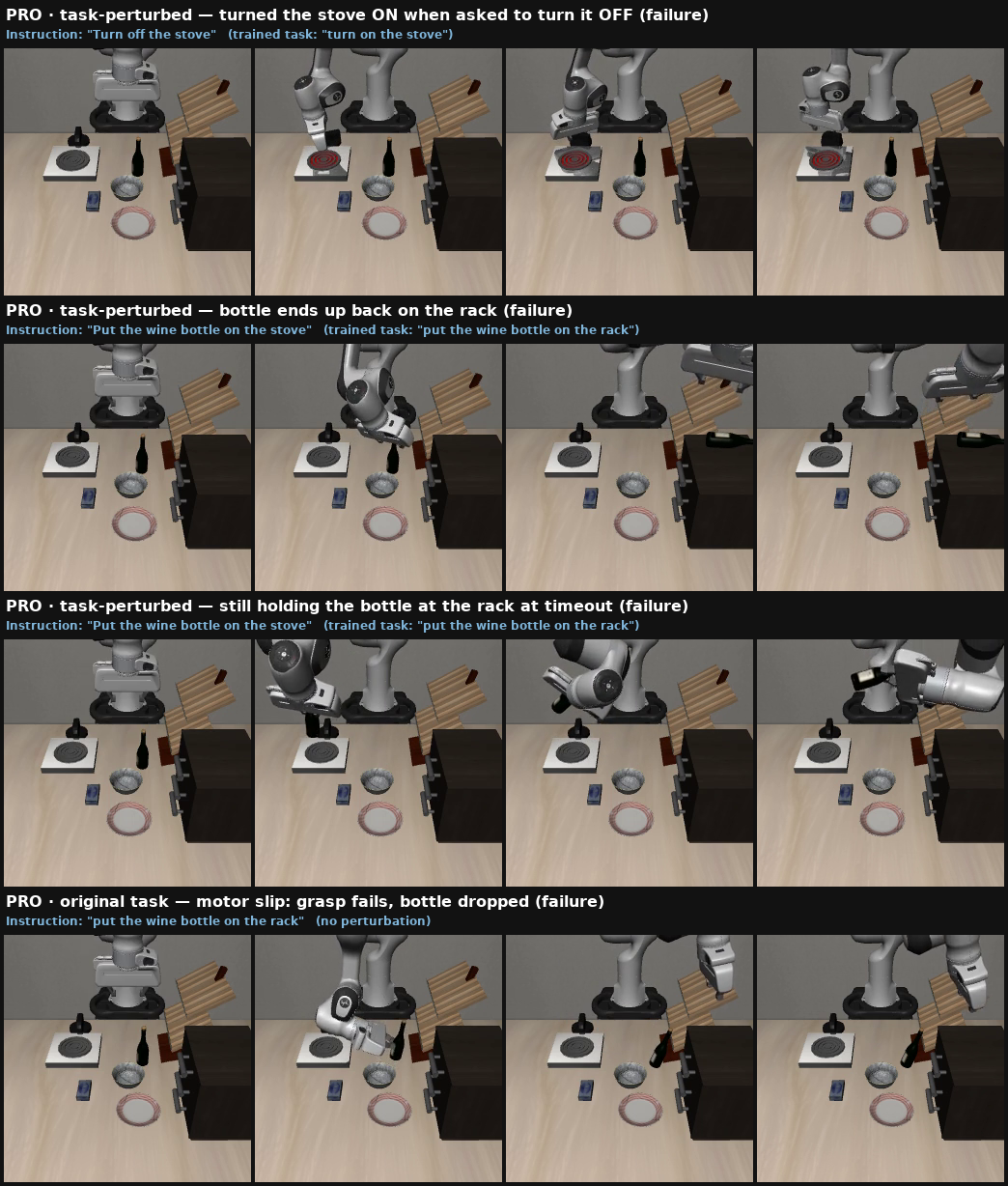}
\caption{Failure analysis on LIBERO-PRO task perturbations (libero\_goal).
Rows 1--3: under task-perturbed instructions, the characteristic failure
is semantic interference---the policy reverts to the behavior of the
trained task. Asked to turn the stove off, it turns the burner on (row 1); asked
to put the wine bottle on the stove, it carries the bottle back to the
trained rack goal (rows 2--3). Row 4: on the unperturbed original task,
failures are motor-level instead (the grasp slips and the bottle is
dropped). Perturbed-instruction failures therefore reflect instruction
generalization rather than manipulation competence. These modes are
systematic rather than cherry-picked: 2/2 recorded failures of ``turn off
the stove'' turn it on, and 2/6 recorded failures of the wine-bottle task
revert to the rack.}
\label{fig:pro-task-failures}
\end{figure*}

\begin{figure*}[t]
\centering
\includegraphics[width=\textwidth]{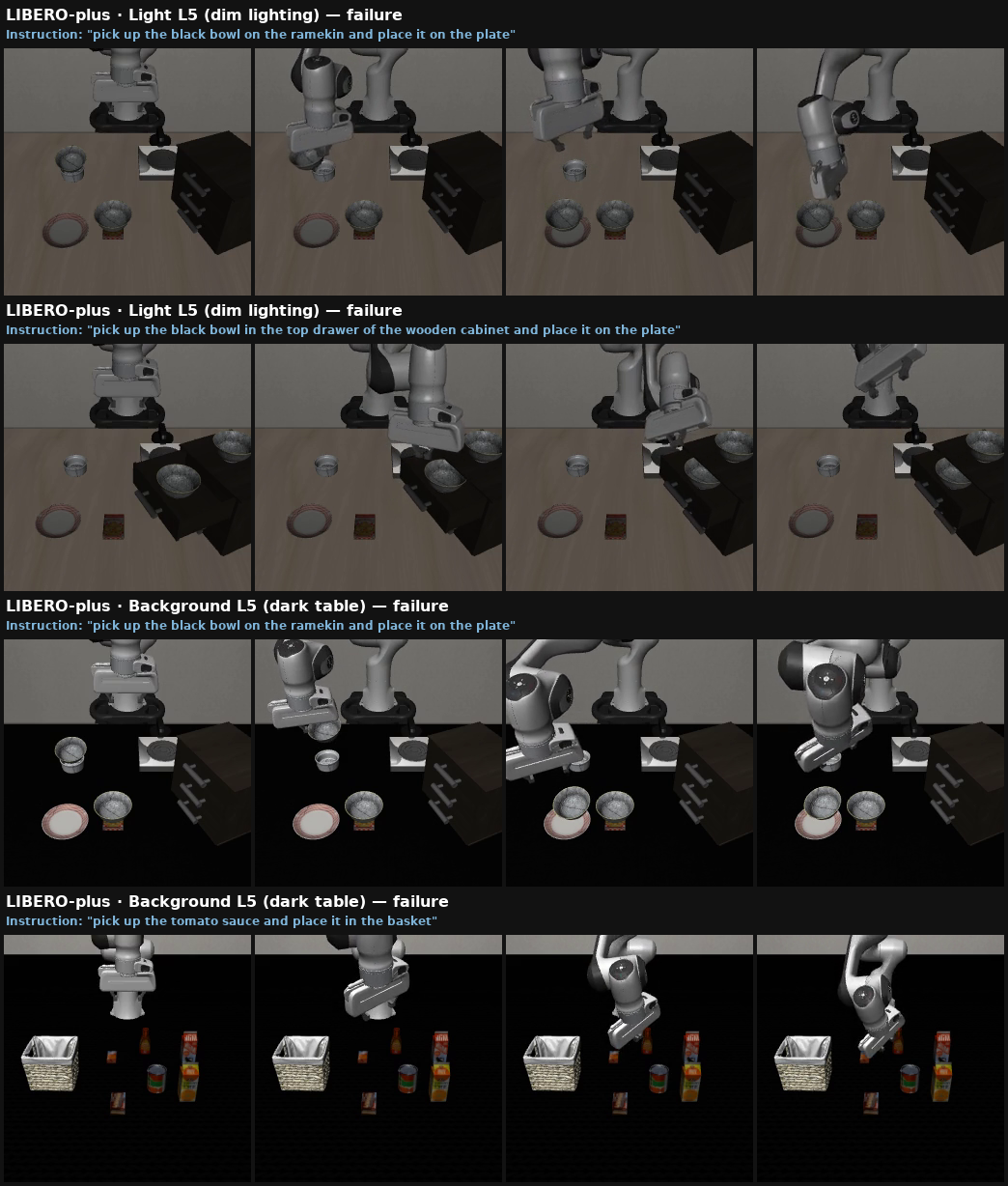}
\caption{Representative failures on LIBERO-plus for the two weakest
perturbation categories of our model: extreme lighting (Light, 53.5\%)
and unseen table textures (Background, 55.6\%), both at difficulty L5.
Rows 1--2 (dim lighting): the policy reaches and manipulates the bowl but
fails to complete the placement---the bowl is knocked beside the plate
(row 1), or the bowl is never extracted from the cabinet (row 2). Rows
3--4 (dark table texture): the policy manipulates the distractor bowl
next to the plate while the target bowl remains on the ramekin (row 3),
and the grasp on the can fails so that it never reaches the basket
(row 4). All episodes run to the step limit without a success.}
\label{fig:plus-failures}
\end{figure*}

This appendix collects qualitative episodes on the two perturbation
benchmarks, complementing the quantitative tables. Figures~\ref{fig:pro-env}
and~\ref{fig:pro-lan} show paired environment- and language-dimension
episodes of LIBERO-PRO---invariance/transfer illustrations, not advantage
claims (the dimensions where the strong baselines lead, Tables~\ref{tab:plus}
and~\ref{tab:pro}). Figure~\ref{fig:pro-task-failures} contrasts
task-perturbed failures with failures on the original task;
Figure~\ref{fig:plus-failures} shows representative failures on the two
weakest LIBERO-plus categories.

\end{document}